\documentclass[journal]{IEEEtran}
\IEEEoverridecommandlockouts
\usepackage{cite}
\usepackage{amsmath,amssymb,amsfonts}
\usepackage{algorithmic}
\usepackage{graphicx}
\usepackage{multirow}
\usepackage{pbox}
\usepackage{textcomp}
\usepackage{hyperref}
\hypersetup{
    colorlinks=true,
    linkcolor=blue,
    filecolor=blue,      
    urlcolor=blue,
    citecolor=blue,
}
\usepackage{xcolor}
\def\BibTeX{{\rm B\kern-.05em{\sc i\kern-.025em b}\kern-.08em
    T\kern-.1667em\lower.7ex\hbox{E}\kern-.125emX}}
\begin{document}

\title{Monolithic Six-DOF Parallel Positioning System for High-precision and Large-range Applications\\
}

\author{\IEEEauthorblockN{Mohammadali Ghafarian\textsuperscript{1,2}, Bijan Shirinzadeh\textsuperscript{1}, Ammar Al-Jodah\textsuperscript{1}} \\
\IEEEauthorblockA{\textsuperscript{1}\textit{Robotics and Mechatronics Research Laboratory (RMRL)} \\
\textit{Department of Mechanical and Aerospace Engineering, Monash University} \\
\textit{Melbourne, VIC 3800, Australia} \\
\textsuperscript{2}\textit{Institute for Intelligent Systems Research and Innovation (IISRI)} \\
\textit{Deakin University}\\
\textit{Waurn Ponds VIC 3216, Australia} \\
m.ghafarian@deakin.edu.au} \\
}

\maketitle

\begin{abstract}
A compact large-range six-degrees-of-freedom (six-DOF) parallel positioning system with high resolution, high resonant frequency, and high repeatability was proposed. It mainly consists of three identical kinematic sections. Each kinematic section consists of two identical displacement amplification and guiding mechanisms, which are finally connected to a limb. Each limb was designed with a universal joint at each end and connected to a moving stage. A computational model of the positioner was built in the ANSYS software package, hence, the input stiffness, output compliance, range, and modal analysis of the system were found. Furthermore, a monolithic prototype made of Acrylonitrile Butadiene Styrene (ABS) was successfully manufactured by the 3D-printing process. It was actuated and sensed by piezoelectric actuators (PEAs) and capacitive displacement sensors, respectively. Finally, the performances of this proposed positioner were experimentally investigated. The positioning resolution was achieved as 10.5nm × 10.5nm × 15nm × 1.8µrad × 1.3µrad × 0.5µrad. The experimental results validate the behavior and capabilities of the proposed positioning system, and also verify the nanometer-scale spatial positioning accuracy within the overall stroke range. Practical applications of the proposed system can be expanded to pick-and-place manipulation, vibration-canceling in microsurgery/micro-assembly, and collaborative manipulators systems.
\end{abstract}


\vspace{-2.5mm}

\section{Introduction}\label{A}
\IEEEPARstart{C}{}ompliant positioning systems with micro/nano manipulation capability have turned into a spotlight of many different industries such as nano-imprint, cell manipulation, scanning probe microscopes, high-density data recording, optical instruments, measurement systems, and aerospace and defense applications \cite{b1,b2}. Monolithic characteristic is an important feature of a compliant positioning system that must be emphasized due to the advantages that accompanying including working smoothly with high resolution, high accuracy, high repeatability, and high dynamic bandwidth. Moreover, it avoids the deficiencies of non-monolithic positioners such as backlash, wear, and friction \cite{b3}. Monolithic manufacturing can be achieved by utilizing advanced technologies such as wire electro-discharge machining (WEDM) \cite{b4,b5}, 3D-printing \cite{b6,b7,b8,b9}, and wet/dry etching \cite{b10}. However, considering the advanced improvements in the 3D-printing manufacturing techniques, spatial designs can be manufactured conveniently using different materials, low manufacturing cost, low scrap rate, and less consumed time.

From the structural point of view, compliant positioning systems can be arranged into three categories, serial \cite{b11}, parallel \cite{b12}, and hybrid \cite{b13,b14} systems. Serial positioners consist of a series of joints connecting the base to the end effector and they suffer from the lack of stiffness and relatively large positioning errors due to their cantilever type of kinematic arrangement. Conversely, parallel positioners offer the advantages of high stiffness, precision, speed capability, and compactness. Despite these characteristics, they have a more complex mechanical design and control algorithms. On the other hand, parallel positioners are composed of a set of parallel kinematic parts with active and passive joints required to maintain the system's mobility and controllability. A combination of both schemes builds up a hybrid positioner. 

Compliant positioning systems are driven by different kinds of actuation principles and they play an important role in the design and applications of these kinds of systems. Electromagnetic actuator (EMA) \cite{b15,b16,b17}, electrostatic actuator (ESA) \cite{b18}, electrothermal actuator (ETA) \cite{b19,b20}, shaped memory alloy (SMA) \cite{b21}, and piezoelectric actuator (PEA) \cite{b22,b23} are among famous actuation principles that have been utilized in the compliant positioning systems. Among the abovementioned principles, the PEA has been widely used due to offering many important advantages including extremely fine resolution, large force generation, fast response, high bandwidth frequency, high electrical-mechanical power conversion efficiency, and small size.

In light of the abovementioned introduction, many studies have focused on introducing new compliant positioners capable of micro/nano manipulation tasks in different working axes. It is evident that the more working axes the positioner possesses, the more difficulties and challenges should be overcome to achieve a controllable system with high precision and accuracy. Therefore, our attention is pointed towards the previous six-DOF positioning systems \cite{b24}. The development of a non-monolithic six-DOF compliant system employing PEAs was presented \cite{b25} which was able to achieve the motion range of 77.42$\mu$m, 67.45$\mu$m, 24.56$\mu$m, 930$\mu$rad, 950$\mu$rad, and 3100$\mu$rad. The design and modeling of a six-DOF compliant dual redundant serial-parallel stage were presented \cite{b26}. The mechanism utilized two different kinds of actuation principles (16 actuators in total), PEA and ultrasonic motor (USM). Therefore, two working modes, micro-motion and macro-motion were captured with the reachable workspaces of 140$\mu$m $\times$ 140$\mu$m $\times$ 135$\mu$m, and 9.7mm $\times$ 9.7mm $\times$ 9.5mm, respectively. Implementing topology optimization, the model of a six-DOF spatial compliant monolithic system with a sub-micron workspace was constructed \cite{b27}, which had the same differential kinematic characteristics as the Gough–Stewart prototype platform. A small range non-monolithic six-DOF micropositioner based on a compliant mechanism with the overall dimension of 241 $\times$ 241 $\times$ 67mm$^3$, and driving with 8 PEAs was introduced \cite{b28}. Considering more PEAs than the number of working axes made the control design of the positioner complicated. The design and screw-based Jacobian analysis of a six-DOF compliant parallel positioning system were investigated \cite{b29}, which was featured by small rotational and out-of-plane translational motions. The design and test of a six-DOF compliant piezo-driven micropositioning system which consisted of two parallel three-DOF compliant positioners assembled serially together were presented \cite{b30}. The proposed positioning system exhibited small translational and rotational workspaces of 8.2$\mu$m, 10.5$\mu$m, 13.0$\mu$m, and 105$\mu$rad, 97$\mu$rad, 224$\mu$rad, respectively. A six-DOF magnetic levitation stage actuated by 8 voice coil motors (VCMs) was presented \cite{b31}. The system was considered to have two upper and lower plates manufactured from aluminum and has an overall dimension of 450 $\times$ 450mm$^2$. The bandwidth frequency and subsequently dynamic characteristics of such a system would be considered to be low due to the levitation characteristic.

According to the above literature review, difficulties associated with the designing and experimentation of six-DOF micropositioning systems including overall size, over-actuation, low resolution, low dynamic characteristic, small working range, measurement strategy, and control technique are the challenges that cause the lack of in-depth experimental studies in this subject. To overcome this need, in-depth computational and experimental investigations of the performance of a compact monolithic large-range six-DOF parallel positioner with high resolution, high resonant frequency, and high repeatability were proposed. Due to the well-established static and dynamic advantages of monolithic structures, the design was manufactured as one piece using a 3D-printing technique. Six PEAs and six high-resolution capacitive displacement sensors were used to generate fine motion and enable capturing a very low resolution, respectively. Static and dynamic characteristics of the six-DOF positioner were investigated utilizing FEA software, ANSYS, and the lowest bandwidth frequency of 137.41Hz and a large working range of 403.7$\mu$m $\times$ 398.5$\mu$m $\times$ 390.94$\mu$m and 8864.4$\mu$rad $\times$ 8297.8$\mu$rad $\times$ 15278.2$\mu$rad were reported, respectively. In a feedback control strategy, an extensive experimental study of the six-DOF positioner was conducted to evaluate different aspects of the characteristics of the proposed system. Based on the captured results, the six-DOF positioner proved to have a low resolution, high precision, high bandwidth frequency, and a very low hysteresis characteristic. Based on the findings, the proposed system will be a suitable option to be implemented in applications such as pick-and-place manipulation, carbon nanotube (CNT) harvesting, automated microscope focus, precision beam steering modules, tremor compensation in microsurgery and assembly of micro-machines, and collaborative manipulators system.

\vspace{-3mm}

\section{Mechanical Design}\label{B}
Figs. 1(a-c) show the isometric, top, and side views of the six-DOF positioner, and its spatial dimensions with respect to the X, Y, and Z axes, which are 172.78mm, 160.20mm, and 24.27mm, respectively.\\
As shown in Figs. 1(a-c), six bridge mechanisms and six leaf parallelogram mechanisms are used, and they are connected to the stage with three arms. Each arm consists of two universal joints. One at the very beginning (connected to the leaf parallelogram mechanism) and another at the end (connected to the stage). The arms are positioned at an equal angle with respect to the horizontal surface (30$^{\circ}$). The bridge mechanism acts as an amplifier for the input PEA, and the leaf parallelogram mechanism allows translational displacement in only one direction. Each universal joint has two DOF and allows two different rotations on two perpendicular surfaces. The universal joints allow the arm to bend at the center point of each joint and preventing the arm to be subjected to an undesired motion such as torsion. Six PEAs can be placed in the middle of the six bridge mechanisms and apply the necessary input forces to the mechanism to generate the desired motions.\\
In the proposed positioner, the dimension of the motion space is $\lambda$ = 6; the number of links is $n$ = 8; the number of joints is $j$ = 9. Using a Kutzbach-Grübler criterion, the DOF of the positioner is calculated as below,

\vspace{-5mm}

\begin{equation}
\begin{split}
M&=\lambda(n-j-1)+\sum_{i=1}^{j} m_{i}=\\
&6\times(8-9-1)+(6\times2+3\times2)=6
\end{split}
\end{equation}

\begin{figure}[htbp]\centering
\begin{tabular}{l}
        \hbox{\hspace{-0.5em}\includegraphics[width=5cm]{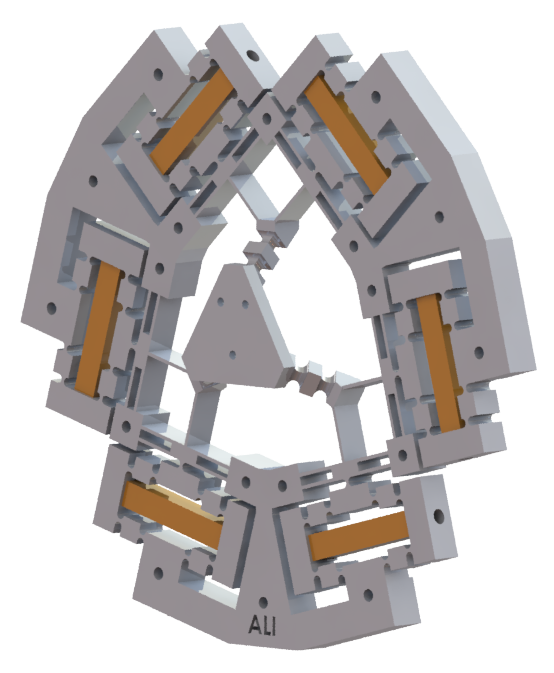}} \\ [0ex]
        \hspace{6em} a \\ [0ex]
        \hbox{\hspace{-0.5em}\includegraphics[width=5cm]{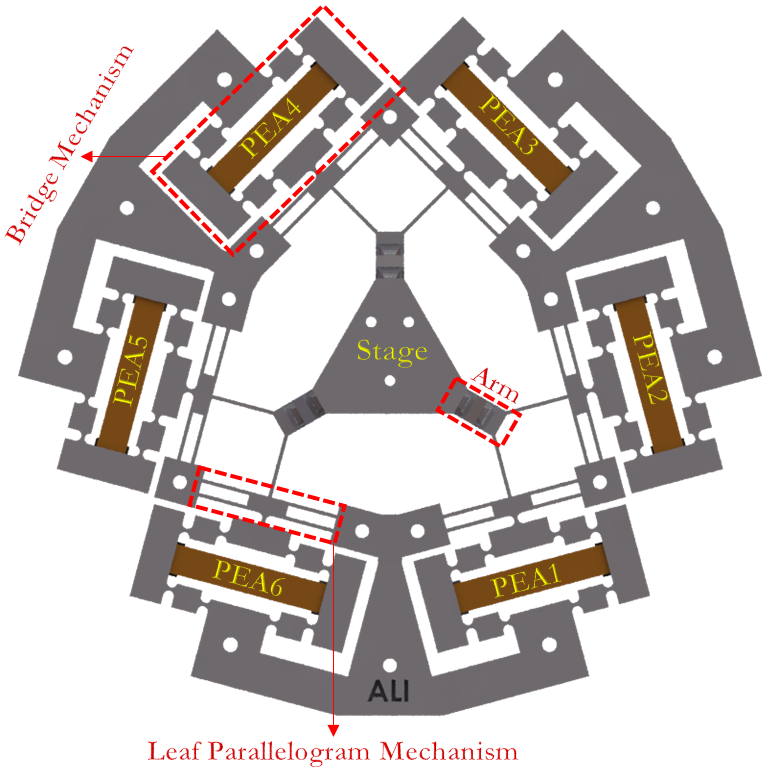}} \\ [0ex]
        \hspace{6em} b \\ [0ex]
        \hbox{\hspace{-0.5em}\includegraphics[width=5cm]{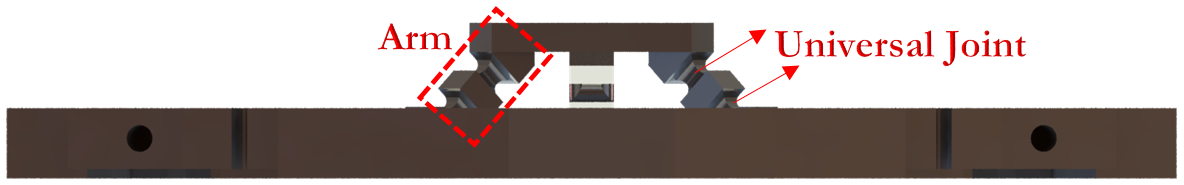}} \\ [0ex]
        \hspace{6em} c \\ [0ex]
	\end{tabular}
\caption{3D solid model of the six-DOF positioner; (a) Isometric view (b) Top view (c) Side view}
\label{fig_1}
\vspace{-3mm}
\end{figure}

\vspace{-3mm}

\section{Computational Analysis}\label{C}
The behavior of the proposed positioner was investigated using simulation software (ANSYS). ABS was considered as the manufacturing material and the mechanical and physical properties were shown in Table 1. During the simulations, 175000 tetrahedral elements (Tet10) were placed in the model. Additionally, the meshing process was performed adaptively meaning sensitive areas such as hinges were filled with a smaller element size (more number of elements). The average aspect ratio (AR) of the elements was determined to be 2.24 which according to the finite element method (FEM) guideline \cite{b32} suggests that ARs should not exceed 3 to obtain good quality meshing, accurate results, and avoid elements distortion.

\begin{table}[!t]
	\renewcommand{\arraystretch}{1.3}
	\caption{Physical and mechanical properties of ABS material}
	\centering
	\label{table_1}
	\resizebox{\columnwidth}{!}{
		\begin{tabular}{l l l}
			\hline\hline \\[-3mm]
			\multicolumn{1}{l}{\textbf{Symbol}} & \multicolumn{1}{l}{\textbf{Quantity}} & \multicolumn{1}{l}{\pbox{20cm}{\textbf{Value}}}  \\[0ex] \hline
			\pbox{20cm}{$ \mathrm{\nu} $} & \pbox{20cm}{Poisson's ratio} &  \pbox{20cm}{$ 0.35 $ } \\ 
			$ \mathrm{\rho} $  & Density & $ 908.7 \mathrm{(kg/m^3)} $ \\[0ex]
			$ \mathrm{E} $ & Young's modulus & $ 2.2 \mathrm{(GPa)} $ \\[0ex]
			\pbox{20cm}{$ \mathrm{\sigma_{yield}} $ } & \pbox{20cm}{Tensile yield strength } & \pbox{20cm}{$ 31 \mathrm{(MPa)} $ } \\ [0ex]
			\pbox{20cm}{$ \mathrm{\sigma_{ultimate}} $ \\ \hphantom{1}} & \pbox{20cm}{Tensile ultimate strength } & \pbox{20cm}{$ 55 \mathrm{(MPa)} $ } \\  [0ex]
			\hline\hline
		\end{tabular}
	}
	\vspace{-3mm}
\end{table}

\vspace{-2.5mm}

\subsection{Modal Analysis}\label{C1}
The dynamic behavior of the positioner was examined to verify the high bandwidth characteristic and the desired output motions. The first six natural frequencies and their corresponding mode shapes of the system are shown in Figs. 2(a-f). It can be seen that the first natural frequency is 137.41Hz. The proposed polymer-made six-DOF parallel positioner possesses high bandwidth frequency due to monolithic characteristic. Thus, this result confirms the repeatability and stability of the system.

\begin{figure*}[htbp]\centering
\begin{tabular}{l l}
        \hbox{\hspace{-0.5em}\includegraphics[width=7cm]{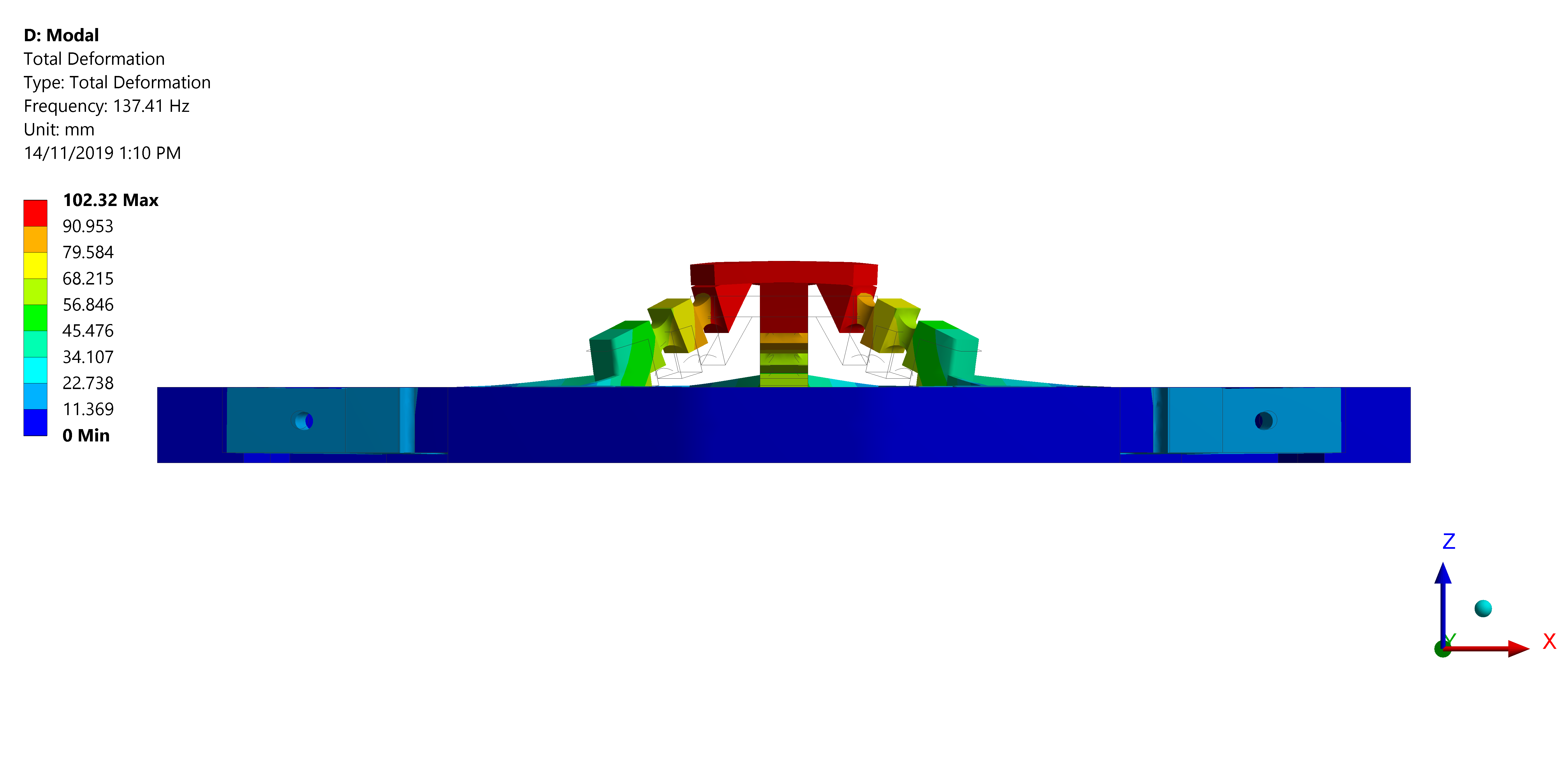}} & \hbox{\hspace{-0.5em}\includegraphics[width=7cm]{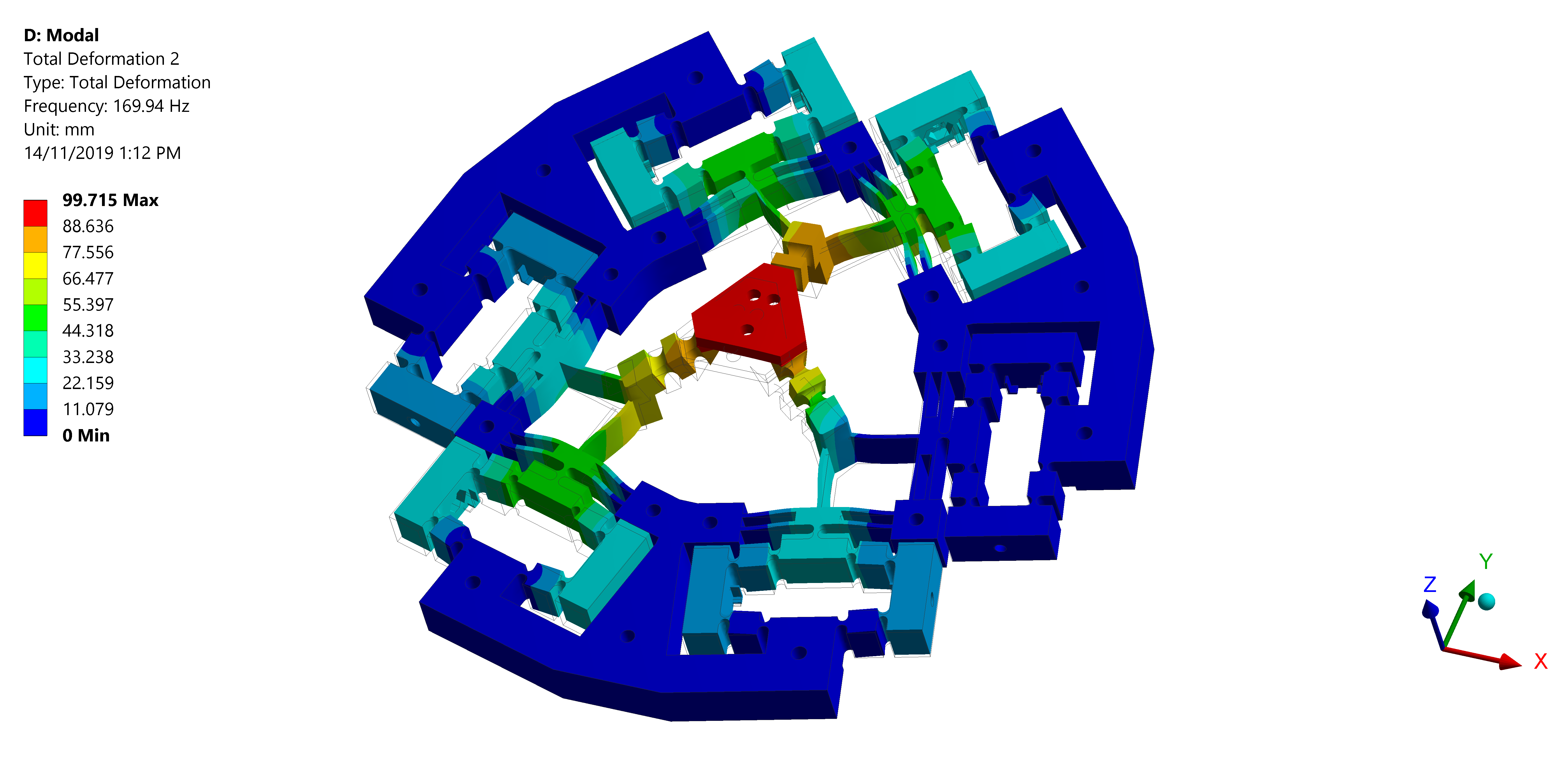}} \\ [0ex]
        \hspace{9em} a & \hspace{9em} b \\ [0ex]
        \hbox{\hspace{-0.5em}\includegraphics[width=7cm]{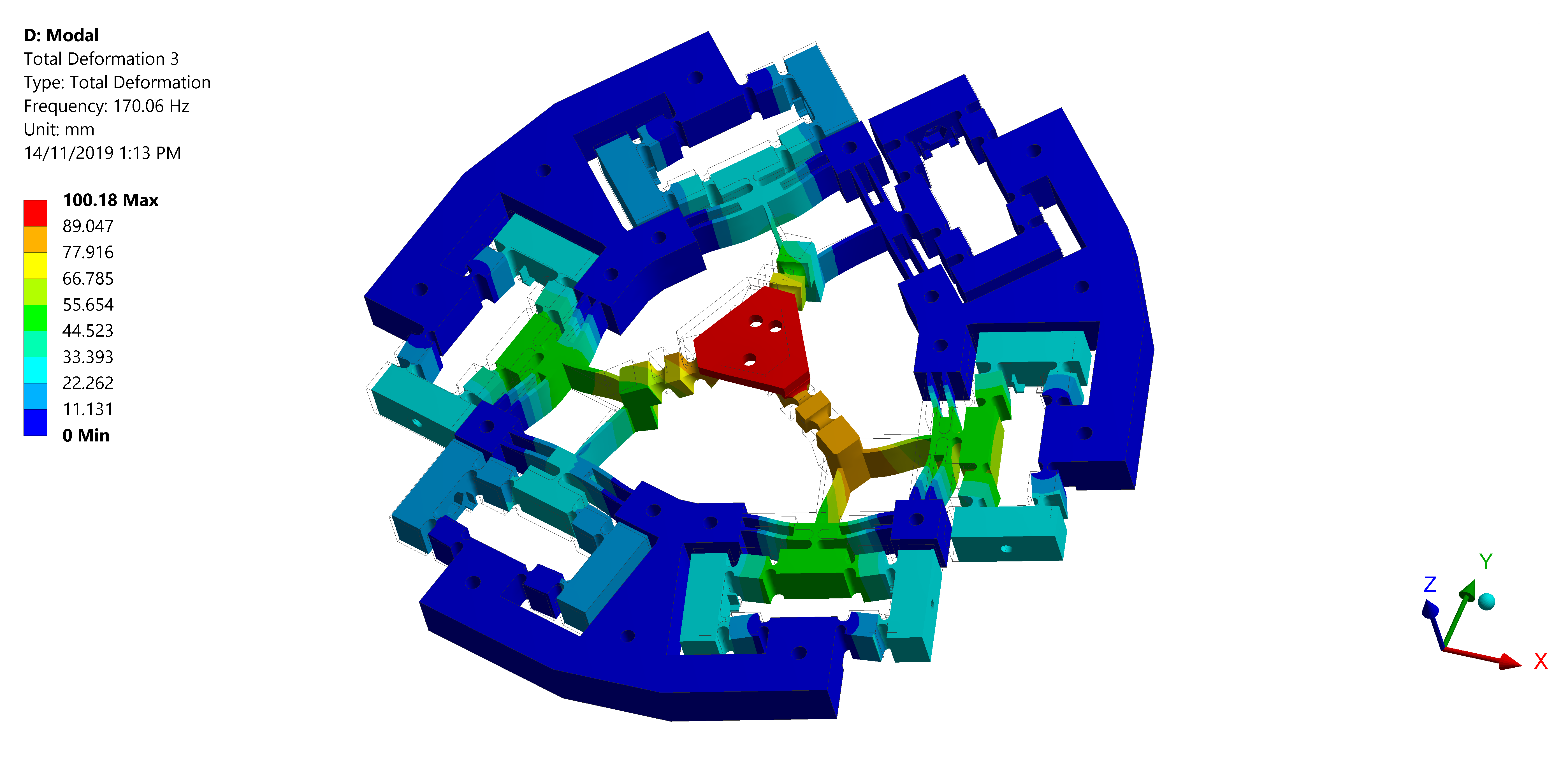}} & \hbox{\hspace{-0.5em}\includegraphics[width=7cm]{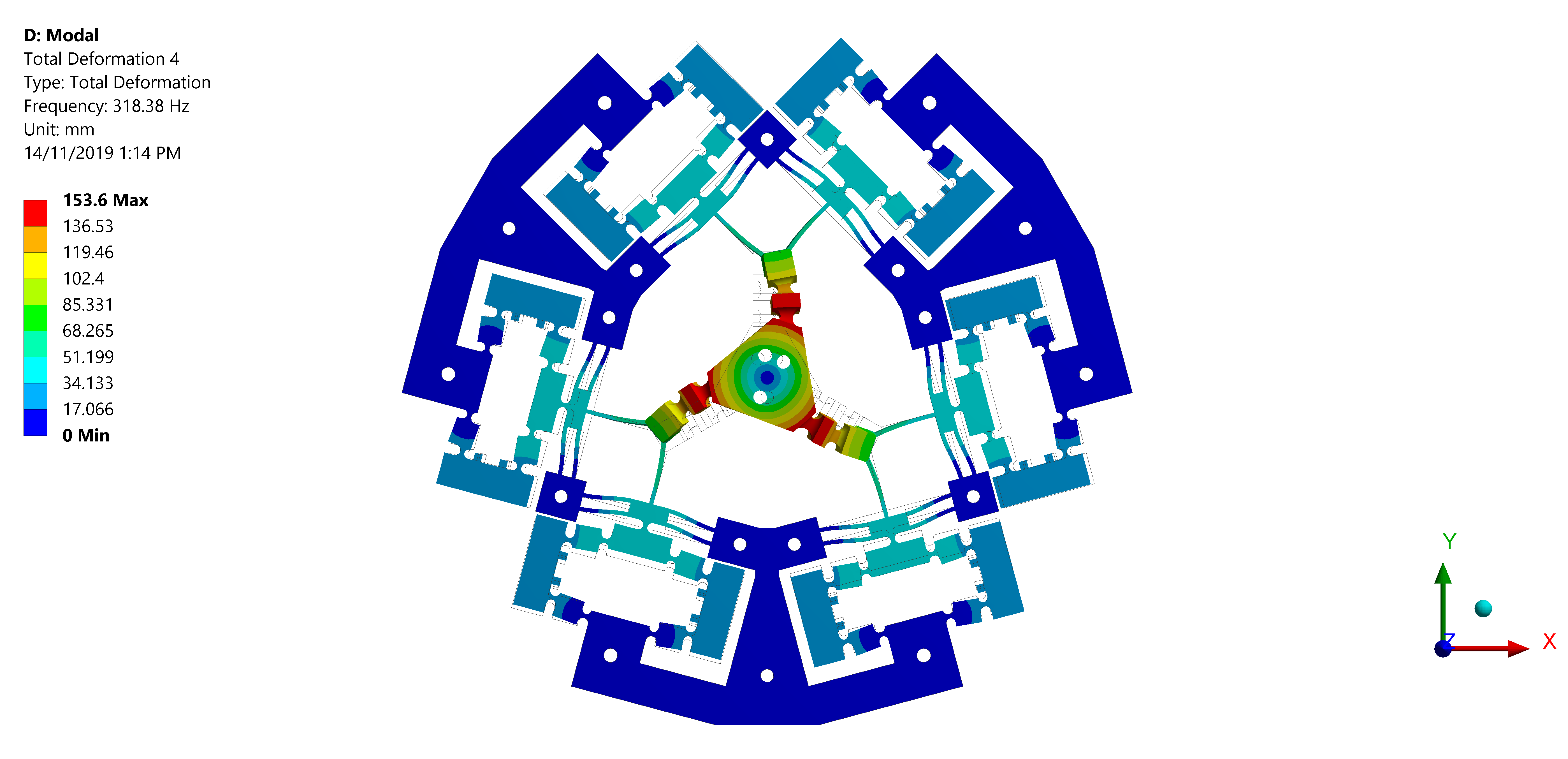}} \\ [0ex]
        \hspace{9em} c & \hspace{9em} d \\ [0ex]
        \hbox{\hspace{-0.5em}\includegraphics[width=7cm]{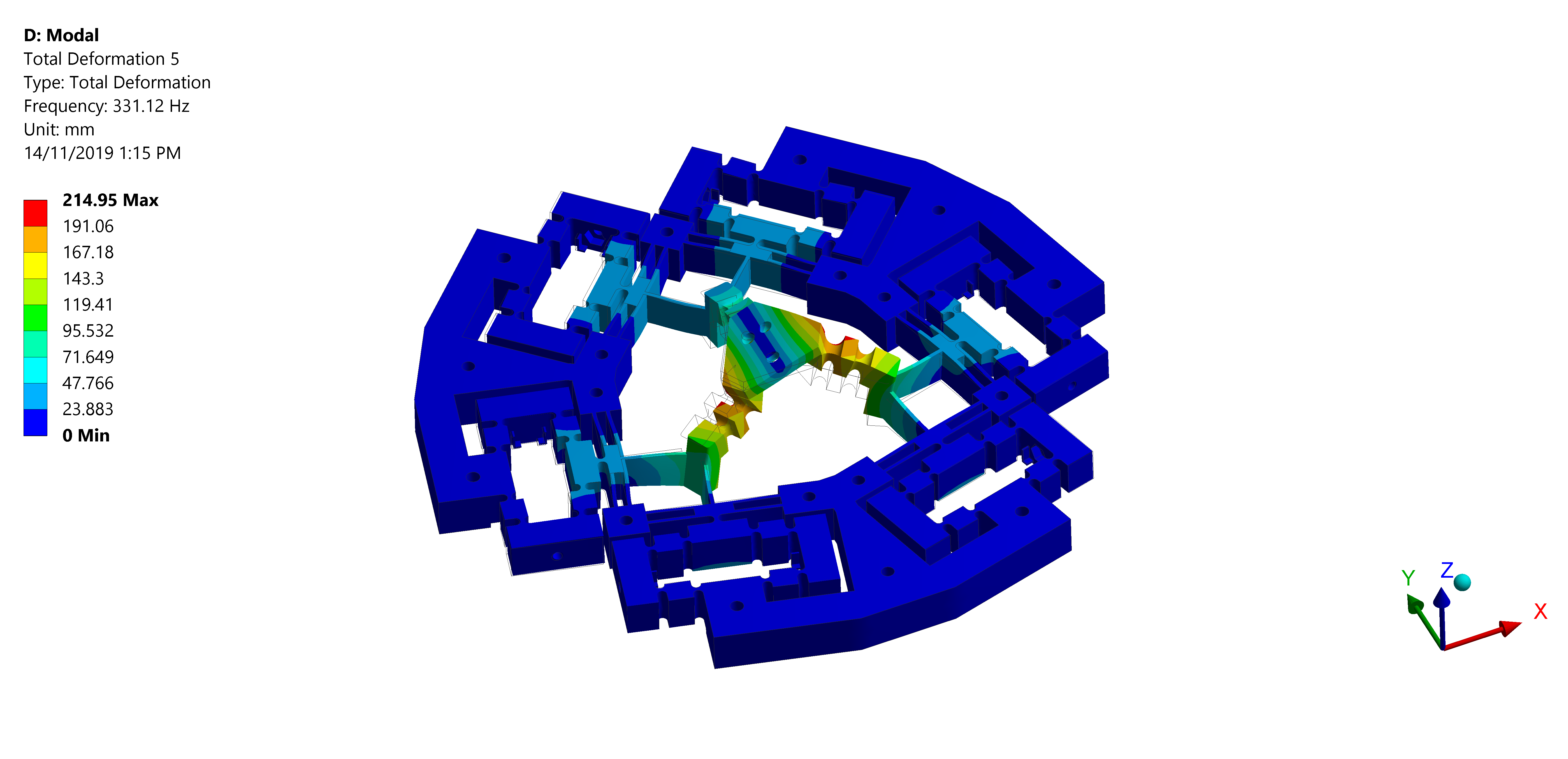}} & \hbox{\hspace{-0.5em}\includegraphics[width=7cm]{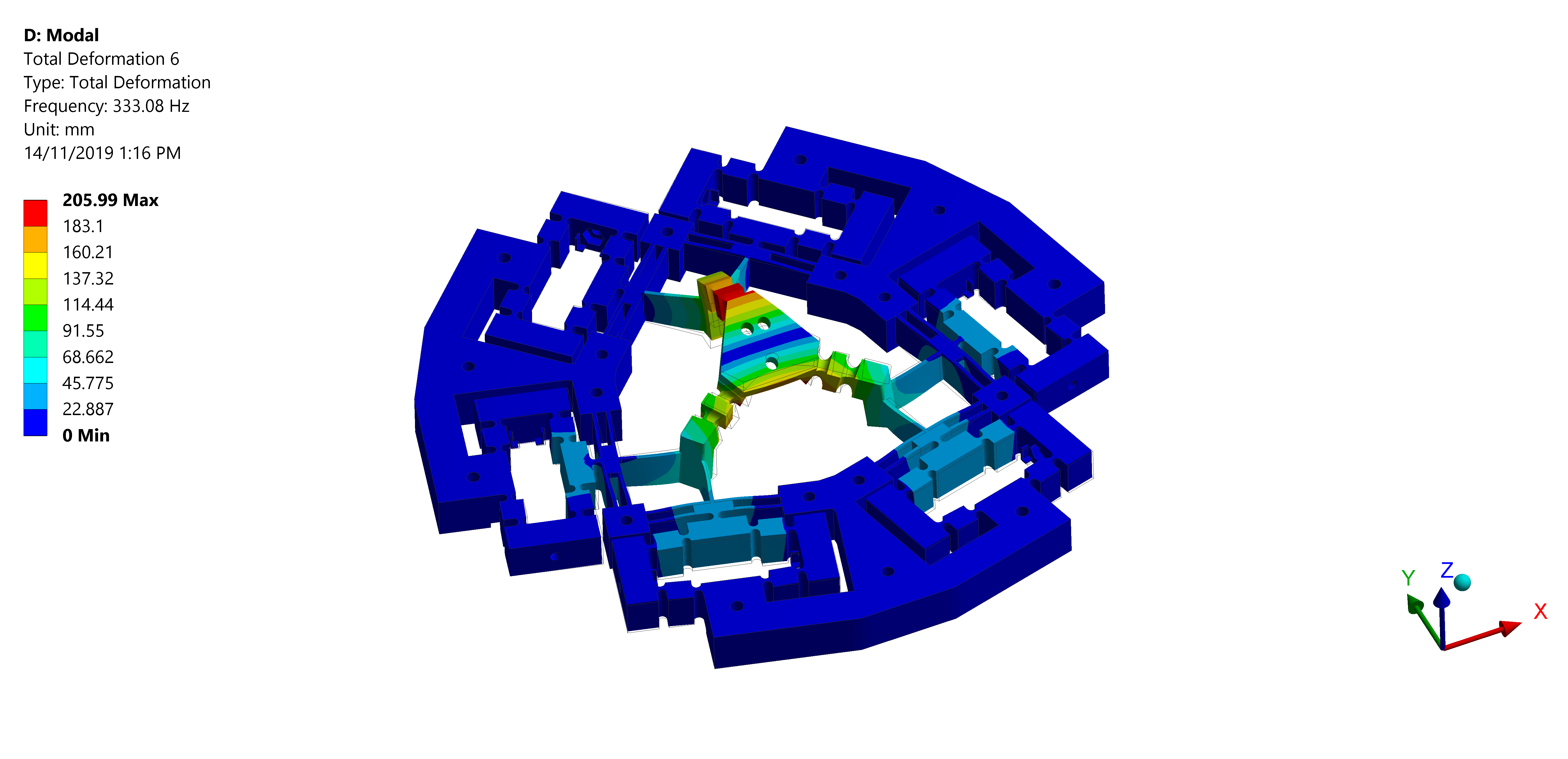}} \\ [0ex]
        \hspace{9em} e & \hspace{9em} f \\ [0ex]
	\end{tabular}
\caption{Mode shapes of the six-DOF parallel positioning system while loaded with 100g apparatus; (a) Z-translation (1st) (b) Y-translation (2nd) (c) X-translation (3rd) (d) $\theta_{z}$-rotation (4th) (e) $\theta_{y}$-rotation (5th) (f) $\theta_{x}$-rotation (6th)}
\label{fig_2}
\vspace{-3mm}
\end{figure*}

\noindent It is evident that adding a mass to the stage of the positioner such as a microgripper or a sensor will cause the system’s natural frequencies to drop. Thus, it is necessary that the designed system possesses a high first natural frequency to maintain its best performance while a tool or an end effector is mounted on the stage. The reported results in Fig. 2 were obtained considering that the positioner carrying 100g apparatus. By removing the apparatus, the proposed positioner possesses a high first natural frequency of 633.25Hz.

\vspace{-1.5mm}

\subsection{Kinematics and Workspace Analysis}\label{C2}
The Jacobian matrix provides the relationship between inputs and outputs of the proposed six-DOF positioning system. This relation is described by Eq. 2,

\vspace{-1mm}

\begin{equation}
\bold{O}_{6\times1}=\bold{J}_{6\times6}\bold{I}_{6\times1}
\end{equation}

\vspace{-1mm}

\noindent where $\bold{O}$ and $\bold{I}$ are the matrices of output and input displacements of the positioner, respectively. The input displacements are generated by six PEAs, and the output displacements are the measured stage’s motions along the six axes using high-resolution displacement sensors.\\
The kinematics of the positioner was calculated by evaluating the stage’s motions across 100 inputs, spanning the full input range, and performing a regression on the resulting outputs. The best kinematics fit is as calculated,

\begin{equation}
\bold{J_{FEA}}=\resizebox{0.7\hsize}{!}{$\begin{pmatrix}
-0.65214 & -0.926 & -0.26136 & 0.26204 & 0.91721 & 0.65117 \\
0.68583 & 0.22332 & -0.90423 & -0.89984 & 0.21785 & 0.69157 \\
0.59421 & 0.59531 & 0.59273 & 0.58933 & 0.58998 & 0.59237 \\
6.8766 & -19.846 & 13.804 & 13.74 & -19.594 & 6.7252 \\
-18.912 & -3.8375 & 14.909 & -14.936 &3.9577 & 18.883 \\
23.248 & -23.313 & 23.035 & -23.009 & 23.089 & -23.199 \\
	\end{pmatrix}$}_{6\times6}
\end{equation}

\noindent It can be noted from Eq. 3 that the motions of the proposed six-DOF parallel positioner are coupled, as the Jacobian matrix is not diagonal. Using this Jacobian, the motion ranges of the positioner along the three translational and three rotational axes will be calculated.
The system’s workspace is dependent on the maximum input displacement that the positioner can endure before it deforms permanently. Therefore, a safety factor analysis had to be investigated to obtain this result. Fig. 3 shows the resultant minimum safety factor of the system. This result was obtained by implementing six equal input displacements to the positioner. It can be noted from Fig. 3 that the minimum safety factor was 1.5 and occurred due to the input displacements of 110µm, which is corresponding to the stress concentration of 20.688MPa (Von-Mises). This stress concentration occurred at one of the universal joints of the positioner (see Fig. 3), which agrees with our expectation. As the universal joints of the system are the most vulnerable part of the designed structure.

\begin{figure}[htbp]\centering
\begin{tabular}{l}
        \hbox{\hspace{-0.5em}\includegraphics[width=7cm]{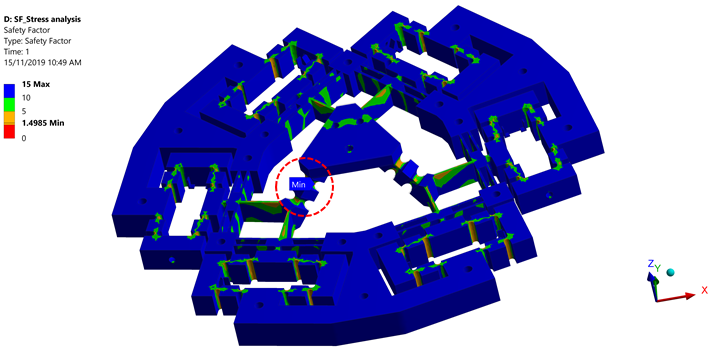}} \\ [0ex]
\end{tabular}
\caption{Safety factor analysis of the proposed positioner}
\label{fig_3}
\vspace{-4mm}
\end{figure}

\noindent The translational and rotational workspaces of the positioner can be calculated using the Jacobian matrix and the information obtained from the safety factor analysis. Thus, by mapping the input and output of the positioner using Eq. 2, the range of motions was found to be 403.7$\mu$m $\times$ 398.5$\mu$m $\times$ 390.94$\mu$m (translational), and 8864.4$\mu$rad $\times$ 8297.8$\mu$rad $\times$ 15278.2$\mu$rad (rotational). Figs. 4(a) and 4(b) illustrate the 3D/2D spaces that are covered by the workspace. In a 3D-space, the volume that is occupied by the translational and rotational motions of the positioner’s end-effector is about 2.0339e+07$\mu$m$^3$ and 3.7015e+11$\mu$rad$^3$, respectively. According to these results, the amplification ratios of the positioner along X, Y, and Z directions are found to be 3.67, 3.62, and 3.55, respectively.

\begin{figure}[t]\centering
\begin{tabular}{l l}
        \hbox{\hspace{-1.5em}\includegraphics[width=5cm]{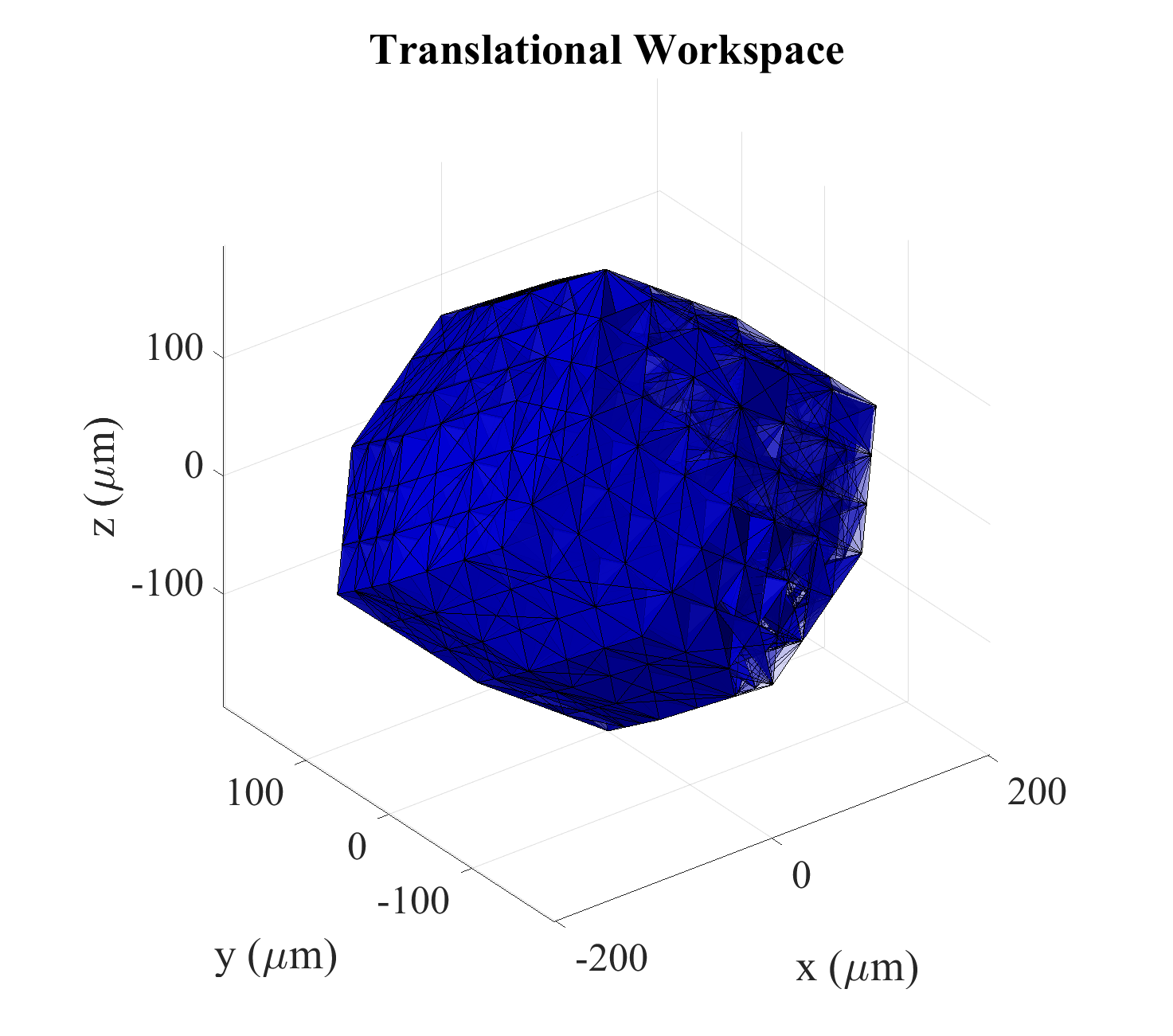}} & \hbox{\hspace{-2.0em}\includegraphics[width=5cm]{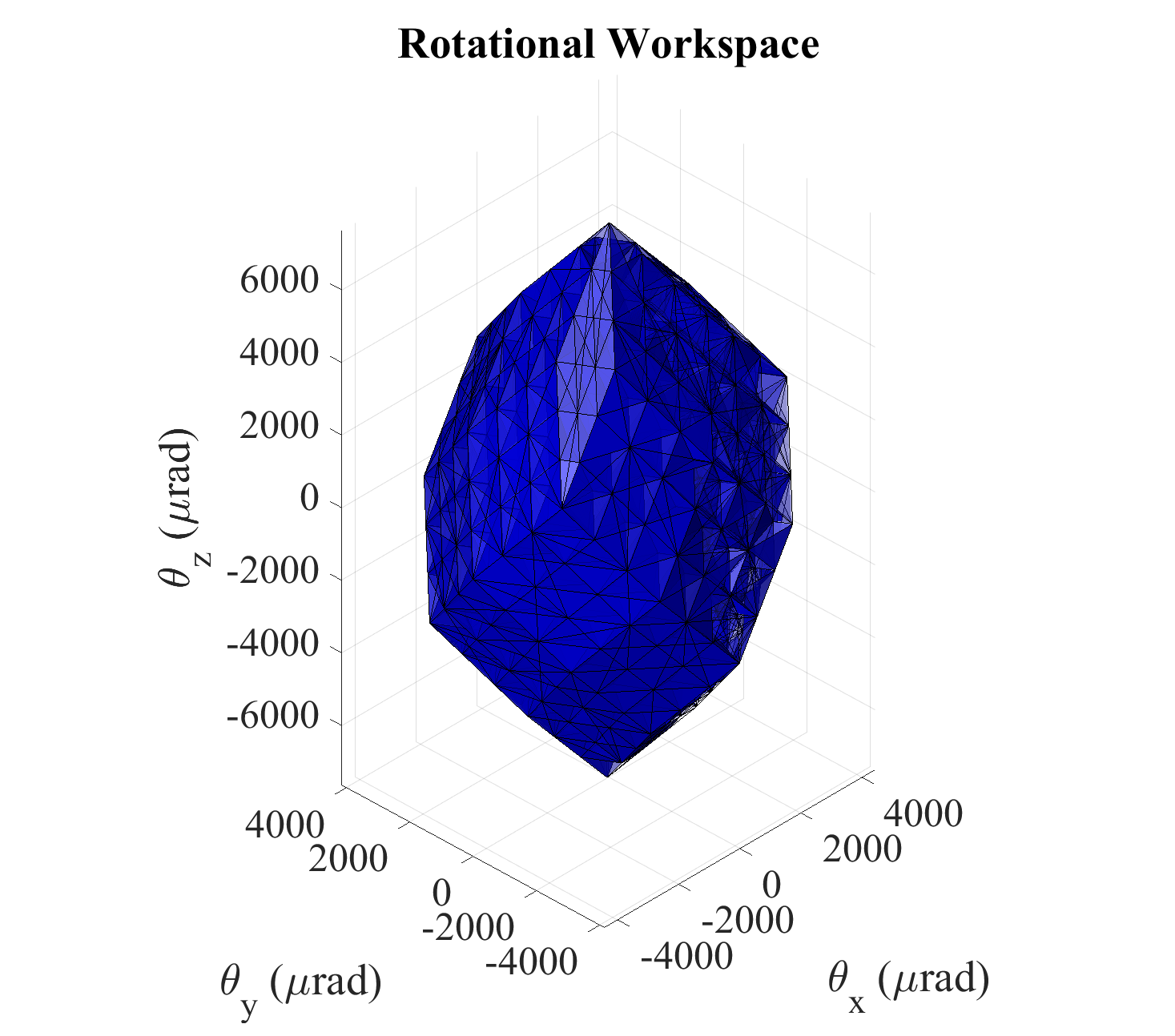}} \\ [0ex]
        \multicolumn{2}{c}{\hspace{0em} a} \\ [0ex]
        \multicolumn{2}{c}{\hbox{\hspace{-0.5em}\includegraphics[width=8cm]{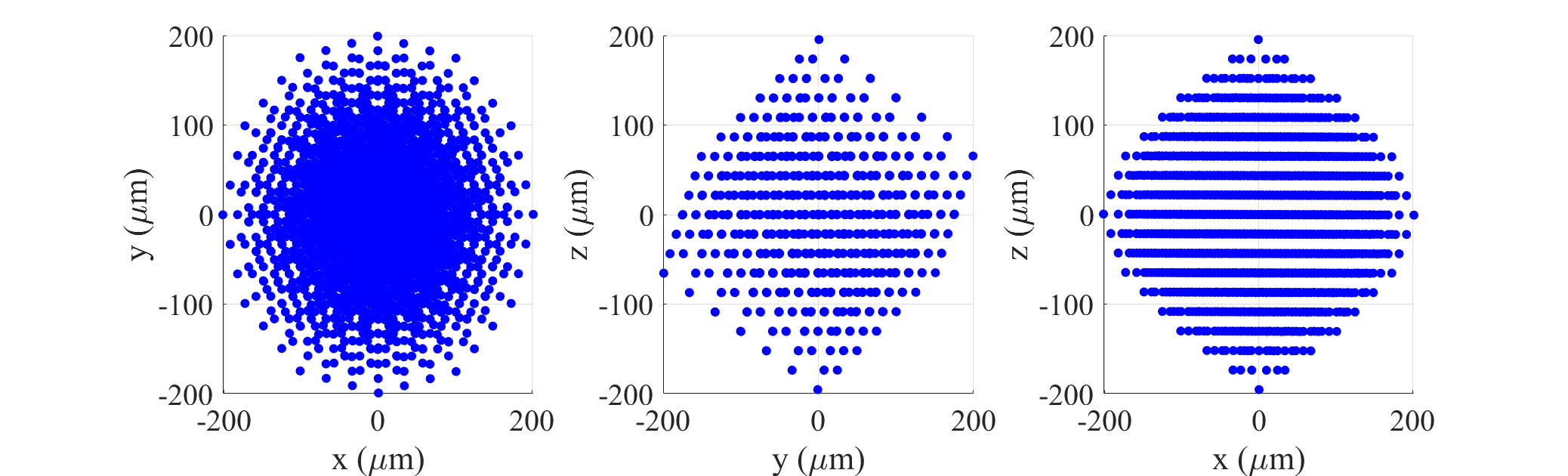}}} \\ [0ex]
        \multicolumn{2}{c}{\hbox{\hspace{-0.5em}\includegraphics[width=8cm]{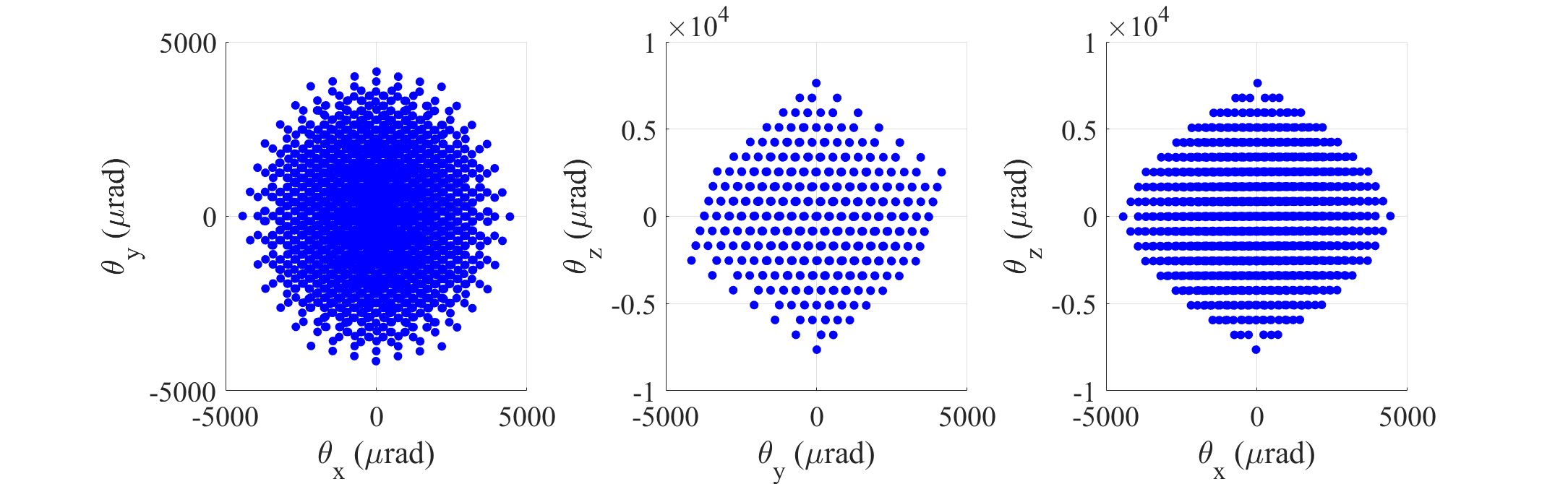}}} \\ [0ex]
        \multicolumn{2}{c}{\hspace{0em} b} \\ [0ex]
	\end{tabular}
\caption{Six-DOF positioner workspace; (a) 3D-View, (b)  Projection of 3D workspace on different working planes}
\label{fig_4}
\vspace{-4mm}
\end{figure}

\vspace{-1.5mm}

\subsection{Input Stiffness and Output Compliance}\label{C3}
In order to calculate the input stiffness, a certain force Fin was applied to the input section of the bridge mechanism. Then, the corresponding input displacement was extracted from the FEA to obtain the input stiffness. In addition, by applying a certain force Fout and moment Mout to the stage, the output compliance can be evaluated by extracting the stage’s displacement and rotation. Based on the simulation results, the input stiffness was found to be 3.0378$\times$10$^6$N/m and the linear and angular output compliances (m/N \& rad/N.m) (Eq. 4, $C_{ij}$=$C_{ji}$ and $C_{ii}>$0) were 8.1702e-06, 8.1760e-06, 10.1760e-06 and 3.0963e-02, 3.0831e-02, 1.8005e-02, respectively. Thus, to reach the full range of the six-DOF positioner, six PEAs with the ability to generate input displacement and force of 110µm and 334.2N, respectively, are essentially needed. Moreover, the linear relationship between the input force and input displacement, Fig. 5, indicates that stress stiffening does not happen in the proposed positioner.

\begin{equation}
\bold{C}=\resizebox{0.75\hsize}{!}{$\begin{pmatrix}
8.1702e-06 & 2.9018e-09 & -8.1048e-09 & 2.2626e-07 & 1.5341e-04 & 6.7613e-07 \\
2.9006e-09 & 8.1760e-06 & 2.1384e-08 & -1.5340e-04 & 1.5135e-07 & -5.8689e-07 \\
-8.1046e-09 & 2.1240e-08 & 1.0176e-05 & -1.2287e-06 & 2.4710e-07 & -2.7619e-07 \\
2.2628e-07 & -1.5340e-04 & -1.2577e-06 & 3.0963e-02 & -2.6427e-07 & 9.1629e-06 \\
1.5341e-04 & 1.5135e-07 & 2.4704e-07 & -2.6427e-07 & 3.0831e-02 & 3.1156e-05 \\
6.9291e-07 & -5.8687e-07 & -2.7620e-07 & 9.1629e-06 & 3.1156e-05 & 1.8005e-02 \\
	\end{pmatrix}$}_{6\times6}
\end{equation}

\vspace{-3mm}

\begin{figure}[htbp]\centering
\begin{tabular}{l}
        \hbox{\hspace{-0.5em}\includegraphics[width=5cm]{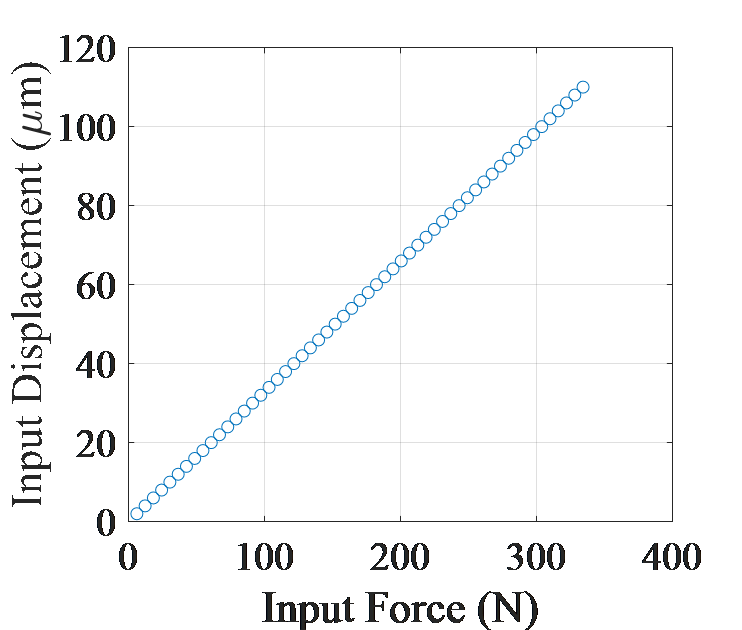}} \\ [0ex]
\end{tabular}
\caption{Relationship between the input force and input displacement of the positioner}
\label{fig_5}
\vspace{-3mm}
\end{figure}

\section{Experimental Verification}\label{D}
The behavior and characteristics of the proposed positioner were evaluated through a series of experimental tests and the results are shown in this section. The developed prototype of the system and the experimental facility are shown in Fig. 6. The basic operation of the positioner was to change the drive voltage from the voltage control unit, THORLABS MDT693A, to the six PEAs from THORLABS (model PK4FYC2) to further drive the positioner. During the installation of the PEAs, pre-compression forces were applied to keep the actuators’ tip and the mechanism in contact. These PEAs are capable of delivering 38.5µm displacement corresponding to a range of operating voltage from 0V to 150V. The position of the moving stage was measured by six Physik Instrumente capacitive sensors (three D-050 and three D-100). The capacitive sensors had a circular active area with a 4.64mm and 6.00mm radius surrounded by an annular guard ring, with a specified working range of 50$\mu$m and 100$\mu$m, respectively. The electronic interface to the capacitive sensors (Physik Instrumente E-509.C3A) provided a distance measurement as an analog voltage, which was recorded at the control computer with a 16-bit analog-to-digital converter (ADC). To align the position capacitive sensors for accurate and precise measurement reading, three Elliot Scientific MDE263 XYZ Micropositioners and three THORLABS MS1 single-axis translation stages were used. In order to reduce the external vibration disturbances, all experimental tests were performed on a Newport pneumatic vibration-isolated optical platform.

\begin{figure}[htbp]\centering
\begin{tabular}{l}
        \hbox{\hspace{-0.5em}\includegraphics[width=6cm]{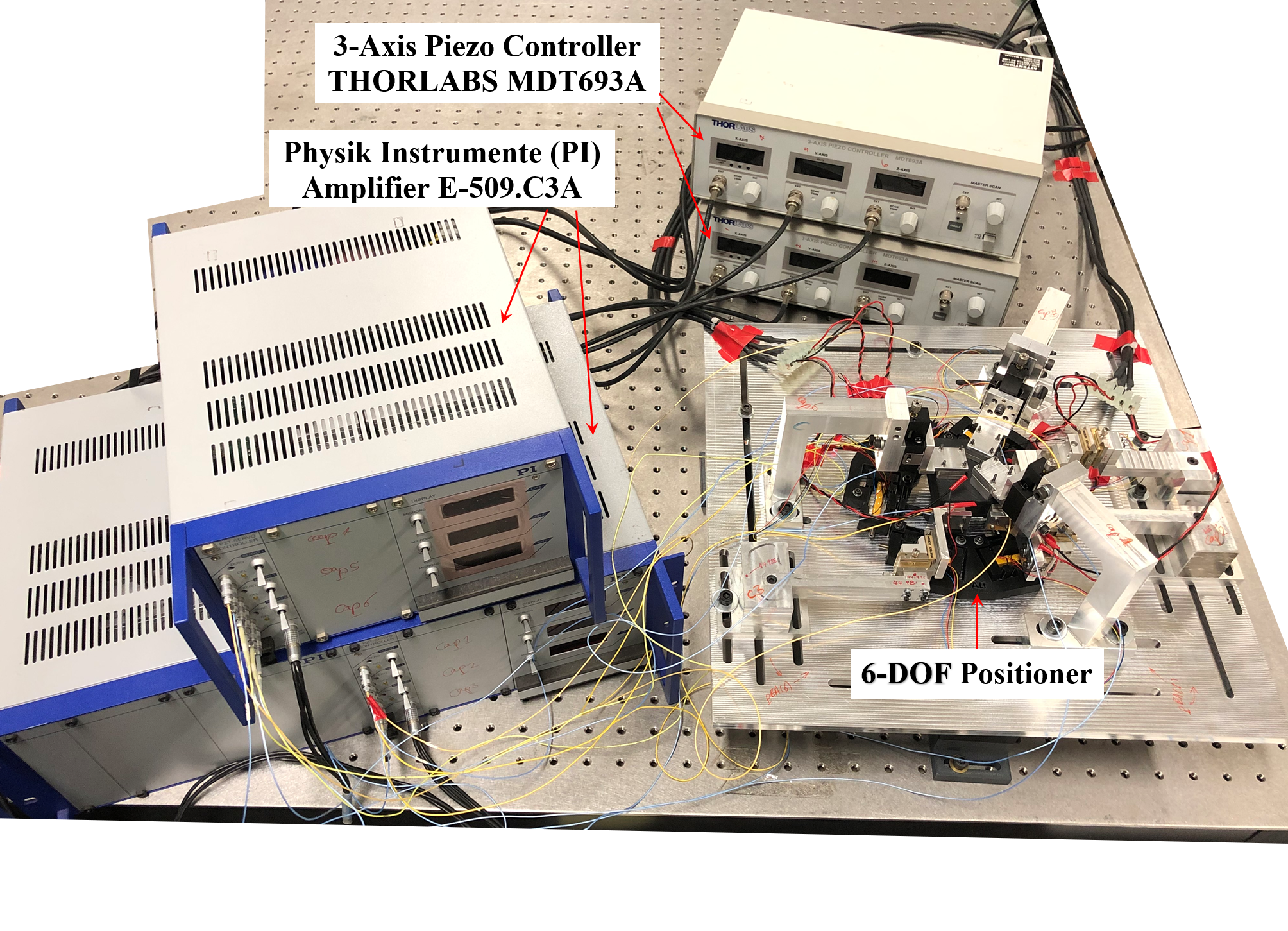}} \\ [0ex]
        \hbox{\hspace{-0.5em}\includegraphics[width=6cm]{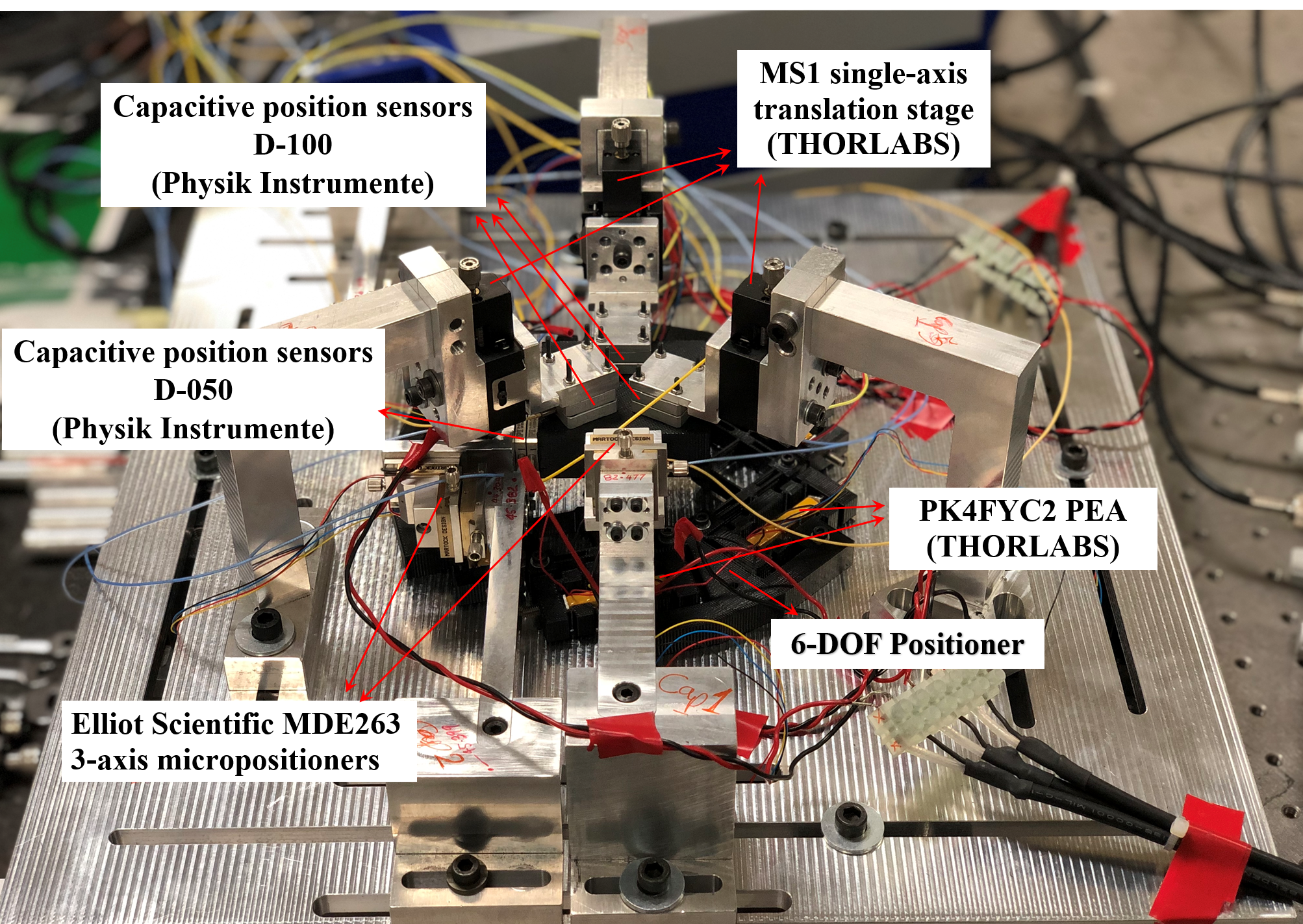}} \\ [0ex]
\end{tabular}
\caption{Experimental setup of the system (six-DOF positioner and equipment)}
\label{fig_6}
\vspace{-3mm}
\end{figure}

\noindent In order to improve the tracking performances and controllability of the positioner, a feedback controller was utilized. Therefore, a Proportional-Integral-Derivative (PID) controller, Eq. 5, was employed to establish feedback for the input signals for precise and controlled positioning tasks. After tunning, the best PID constants were found to be 0.1, 300, and 0 for $k_{p}$ (Proportional Const.), $k_{i}$ (Integral Const.), and $k_{d}$ (Derivative Const.), respectively.

\begin{equation}
u(t) = k_{p}e(t)+k_{i}\int_{0}^{t}e(\tau)d\tau+k_{d}\frac{de(t)}{dt}
\end{equation}

\noindent where $u$ is the control input to the piezo-amplifier modules. Fig. 7 illustrates the schematic diagram of the feedback control strategy implemented in real-time experiments.

\begin{figure}[htbp]\centering
\begin{tabular}{l}
        \hbox{\hspace{-0.5em}\includegraphics[width=8.5cm]{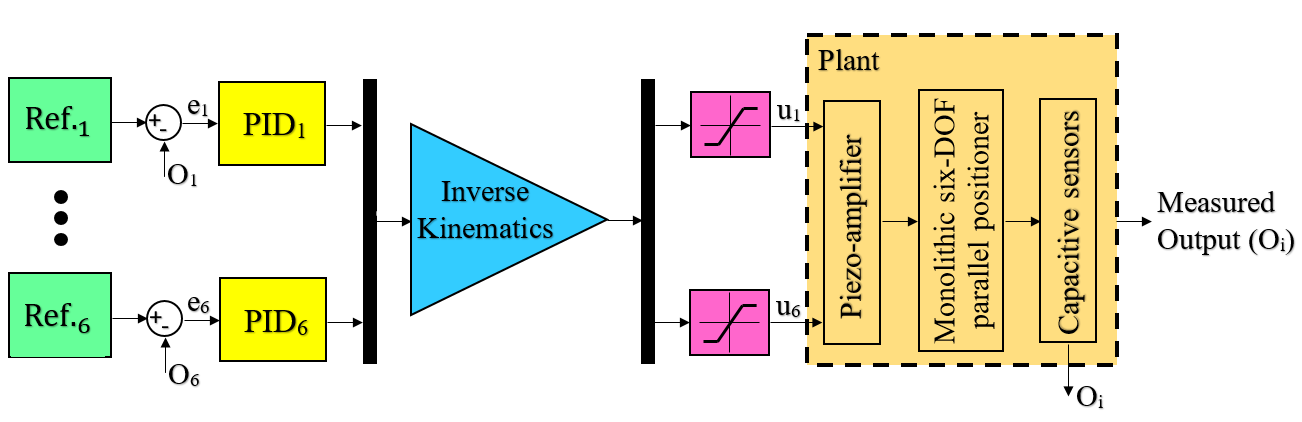}} \\ [0ex]
\end{tabular}
\caption{The schematic diagram of the feedback PID control strategy}
\label{fig_7}
\vspace{-3mm}
\end{figure}

\subsection{Resolution Characterization of the System}\label{D1}
The minimum resolution of the positioner was tested by utilizing stairstep signals with a period of 2$s$. As can be seen in Fig. 8, the minimum resolution was determined for each working axis and found to be 10.5nm along the X-axis, 10.5nm along the Y-axis, 15nm along the Z-axis, 1.8$\mu$rad along the $\theta_{x}$-axis, 1.3$\mu$rad along the $\theta_{y}$-axis, and 0.5$\mu$rad along the $\theta_{z}$-axis, respectively. It is worthy of mention that the resolution of the positioner was influenced by scatter radiation, geometrical effects, and measuring electronic noises. Therefore, the minimum resolution can be improved in the future.

\begin{figure}[htbp]\centering
\begin{tabular}{l l}
        \hbox{\hspace{-1.0em}\includegraphics[width=5cm]{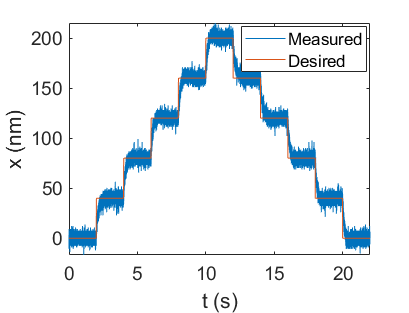}} & \hbox{\hspace{-2.5em}\includegraphics[width=5cm]{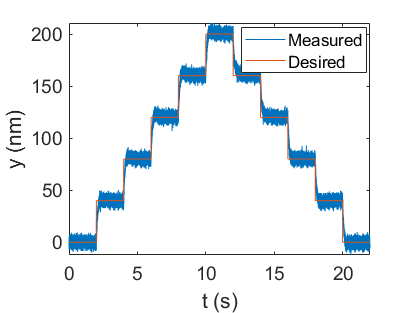}} \\ [0ex]
        \hbox{\hspace{-1.0em}\includegraphics[width=5cm]{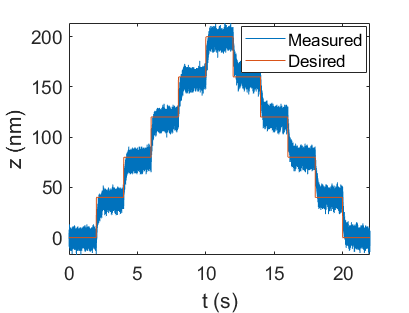}} & \hbox{\hspace{-2.5em}\includegraphics[width=5cm]{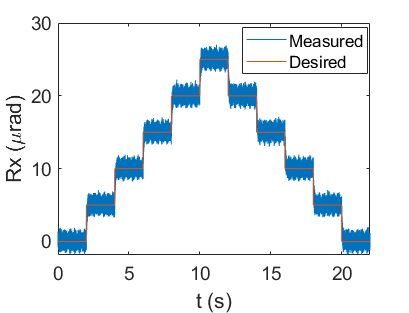}} \\ [0ex]
        \hbox{\hspace{-1.0em}\includegraphics[width=5cm]{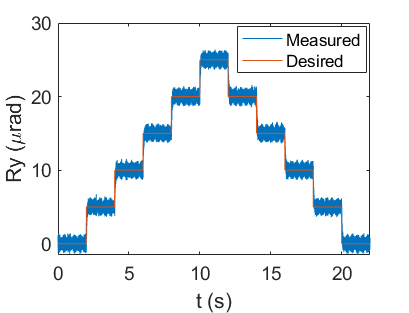}} & \hbox{\hspace{-2.5em}\includegraphics[width=5cm]{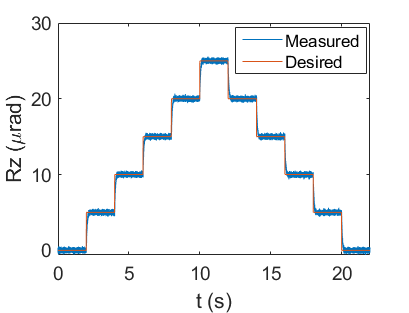}} \\ [0ex]
	\end{tabular}
\caption{System response to applied staircase input}
\label{fig_8}
\vspace{-5mm}
\end{figure}

\vspace{-1.5mm}
\subsection{Motion Tracking and Hysteresis Reduction}\label{D2}
To verify the capability of the developed six-DOF positioner for dynamic motion tracking, a series of 1D, 2D, and 3D complex trajectories were designed and the system was programmed to follow those trajectories. The obtained results are shown in Figs 9-11.\\
In Figs. 9(a,b) and 9(c,d), the positioning paths were two circular and two rose path trajectories, respectively. Each circular or rose path trajectory was designed, so the positioner performs a positioning task in the designated working axes. The corresponding trajectory tracking errors are also shown in Fig. 9.\\
In another experimental testing, the effect of the manipulation range of the positioner on tracking error was examined. Thus, two series of six signals with a constant frequency of 0.5Hz and two different amplitudes (one is two times the other) were sent to the system. According to the captured results in Fig. 10, it can be observed that doubling the manipulation range in all working axes causes an increase in the manipulation inaccuracy, consequently. Furthermore, for some specific applications, it is needed to scan a trajectory with different frequencies. As a result, the effect of manipulation frequency becomes important and provides useful and interesting information regarding the dependency of the accuracy of manipulation on this parameter. Fig. 11 shows the results of experimental testing of trajectory tracking of a similar input signal with the same characteristics such as amplitude and phase, however with different frequencies of 0.1Hz, 0.5Hz, and 1Hz. As it was expected, the results demonstrate the importance of scanning frequency on the accuracy of the manipulation task, and increasing the speed of manipulation leads to more tracking errors. Therefore, there is always a trade-off to be expected.

\begin{figure*}[htbp]\centering
\begin{tabular}{l l l l}
        \hbox{\hspace{-1.5em}\includegraphics[width=5cm]{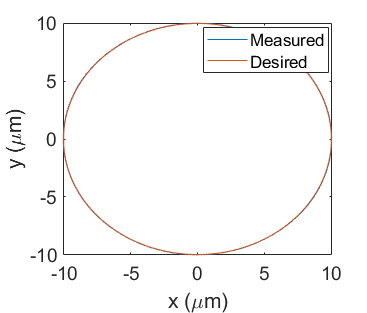}} & \hbox{\hspace{-2.5em}\includegraphics[width=5cm]{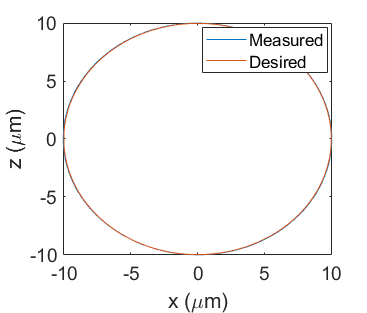}} & \hbox{\hspace{-1.5em}\includegraphics[width=5cm]{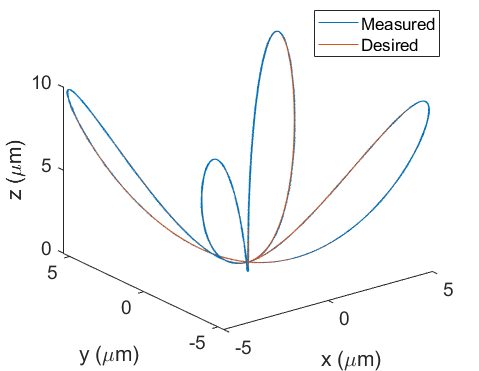}} & \hbox{\hspace{-2.0em}\includegraphics[width=5cm]{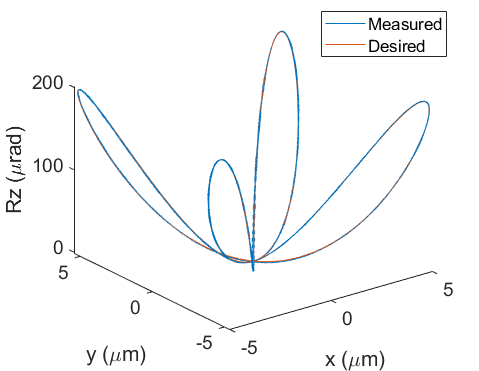}} \\ [0ex]
        \hbox{\hspace{-1.5em}\includegraphics[width=5cm]{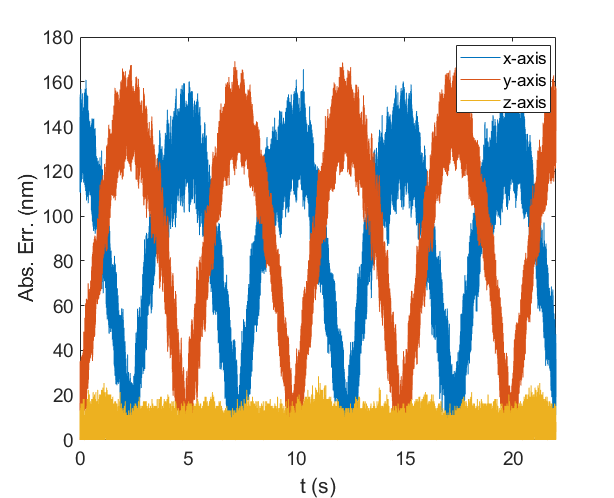}} & \hbox{\hspace{-2.5em}\includegraphics[width=5cm]{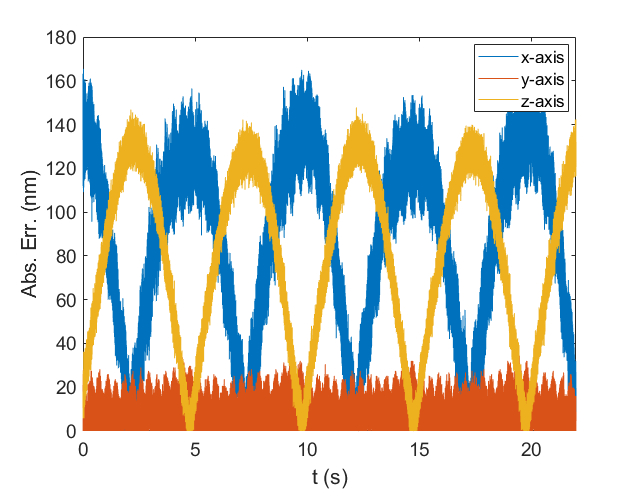}} & \hbox{\hspace{-1.5em}\includegraphics[width=5cm]{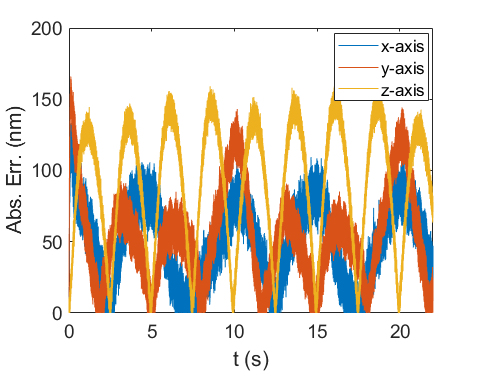}} & \hbox{\hspace{-2.0em}\includegraphics[width=5cm]{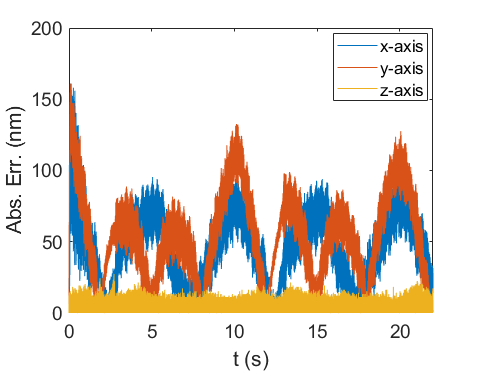}} \\ [0ex]
        \hbox{\hspace{-1.5em}\includegraphics[width=5cm]{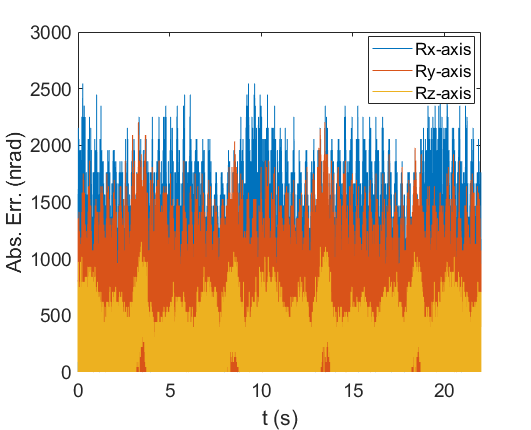}} & \hbox{\hspace{-2.5em}\includegraphics[width=5cm]{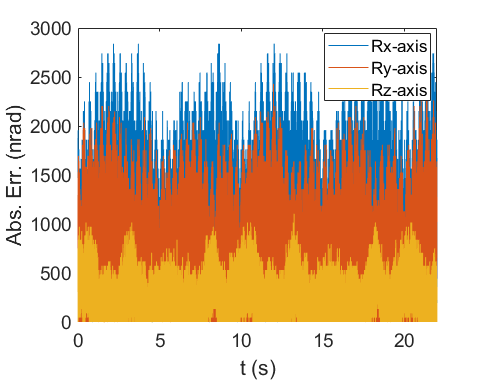}} & \hbox{\hspace{-1.5em}\includegraphics[width=5cm]{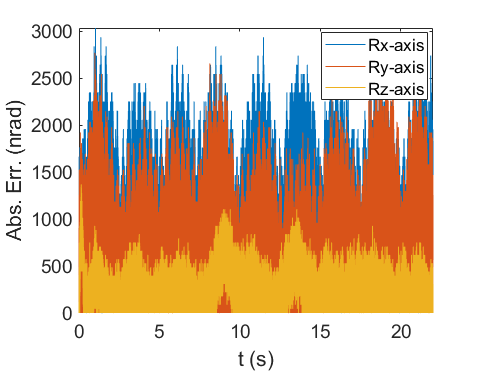}} & \hbox{\hspace{-2.0em}\includegraphics[width=5cm]{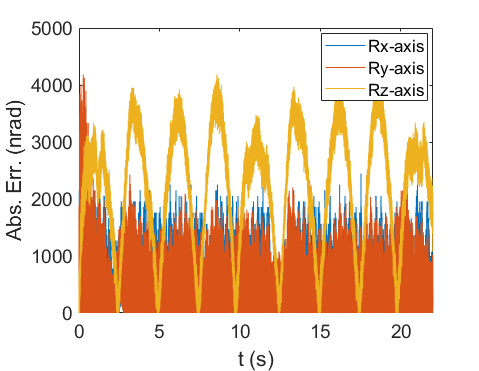}} \\ [0ex]
        \hspace{6em}a & \hspace{6em}b & \hspace{6em}c & \hspace{6em}d\\ [0ex]
	\end{tabular}
\caption{The results of trajectories tracking and their corresponding translational and rotational tracking errors; (a) XY circular trajectory (b) XZ circular trajectory (c) XYZ rose path trajectory (d) XY$\theta_{z}$ rose path trajectory}
\label{fig_9}
\vspace{-3mm}
\end{figure*}

\begin{figure}[htbp]\centering
\begin{tabular}{l l}
        \hbox{\hspace{-1.0em}\includegraphics[width=5cm]{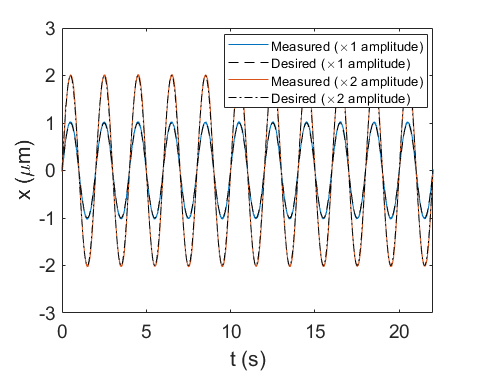}} & \hbox{\hspace{-2.5em}\includegraphics[width=5cm]{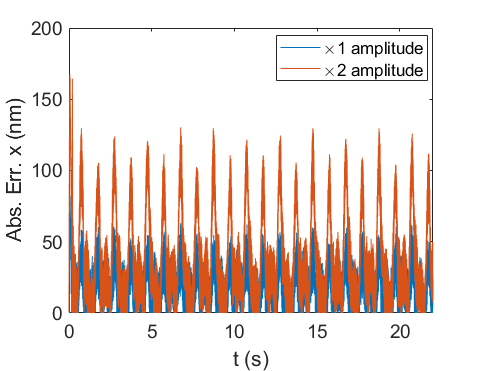}} \\ [0ex]
        \hbox{\hspace{-1.0em}\includegraphics[width=5cm]{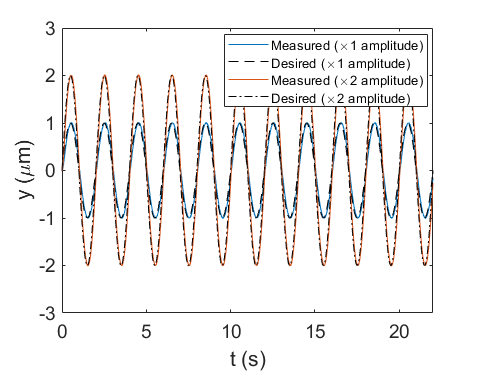}} & \hbox{\hspace{-2.5em}\includegraphics[width=5cm]{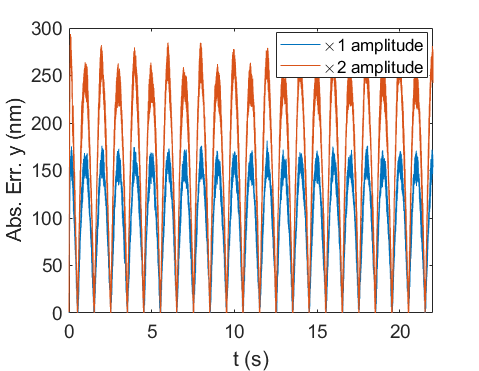}} \\ [0ex]
        \hbox{\hspace{-1.0em}\includegraphics[width=5cm]{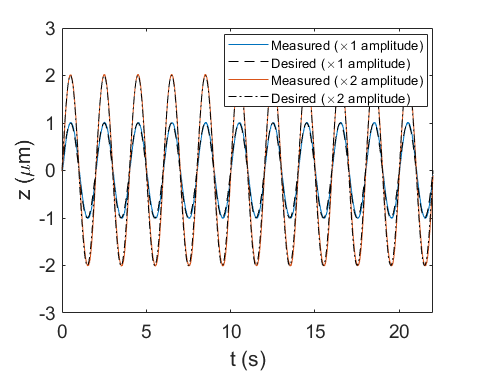}} & \hbox{\hspace{-2.5em}\includegraphics[width=5cm]{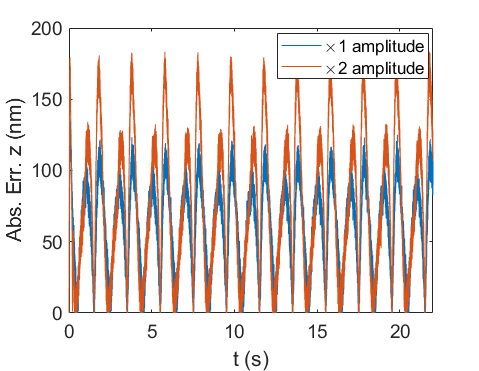}} \\ [0ex]
        \hbox{\hspace{-1.0em}\includegraphics[width=5cm]{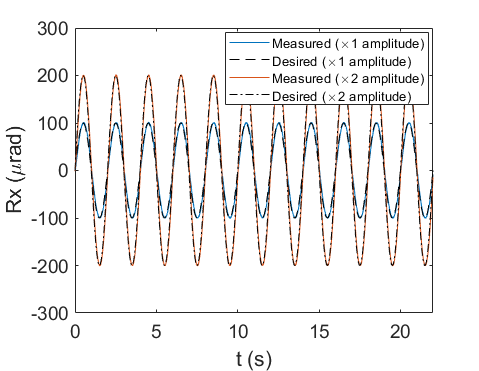}} & \hbox{\hspace{-2.5em}\includegraphics[width=5cm]{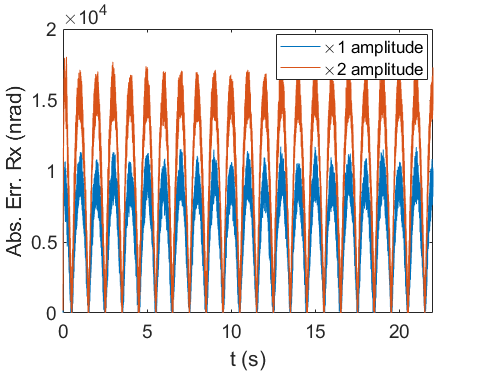}} \\ [0ex]
        \hbox{\hspace{-1.0em}\includegraphics[width=5cm]{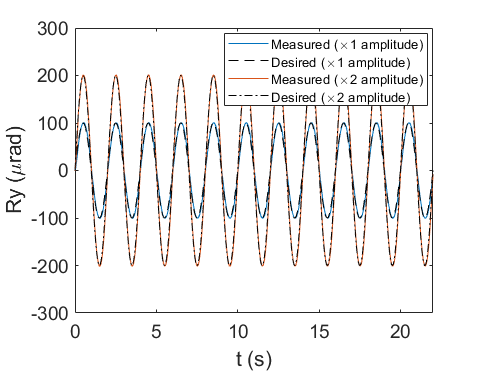}} & \hbox{\hspace{-2.5em}\includegraphics[width=5cm]{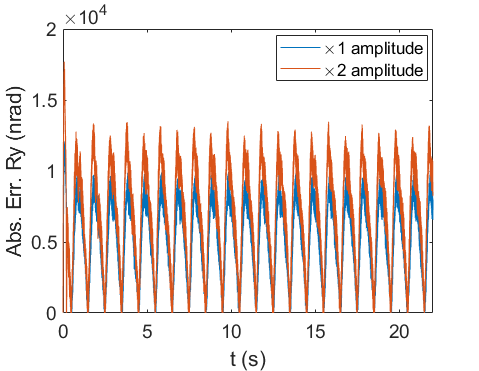}} \\ [0ex]
        \hbox{\hspace{-1.0em}\includegraphics[width=5cm]{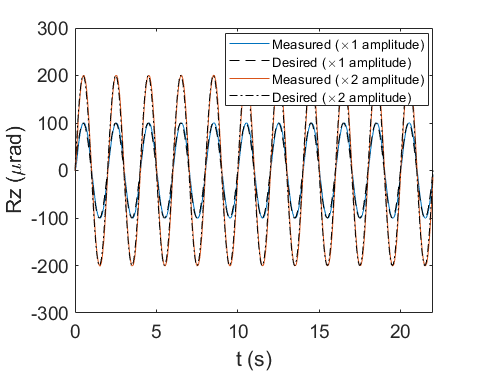}} & \hbox{\hspace{-2.5em}\includegraphics[width=5cm]{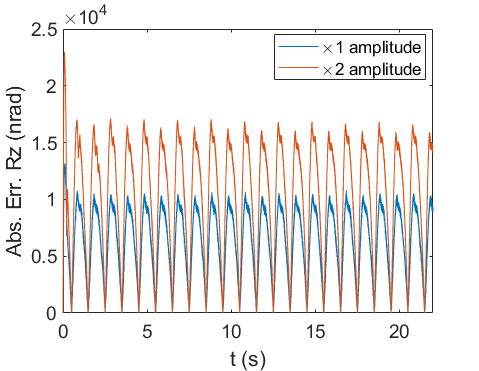}} \\ [0ex]
	\end{tabular}
\caption{Effect of manipulation range on the accuracy of the manipulation task}
\label{fig_10}
\end{figure}

\begin{figure}[htbp]\centering
\begin{tabular}{l l}
        \hbox{\hspace{-1.0em}\includegraphics[width=5cm]{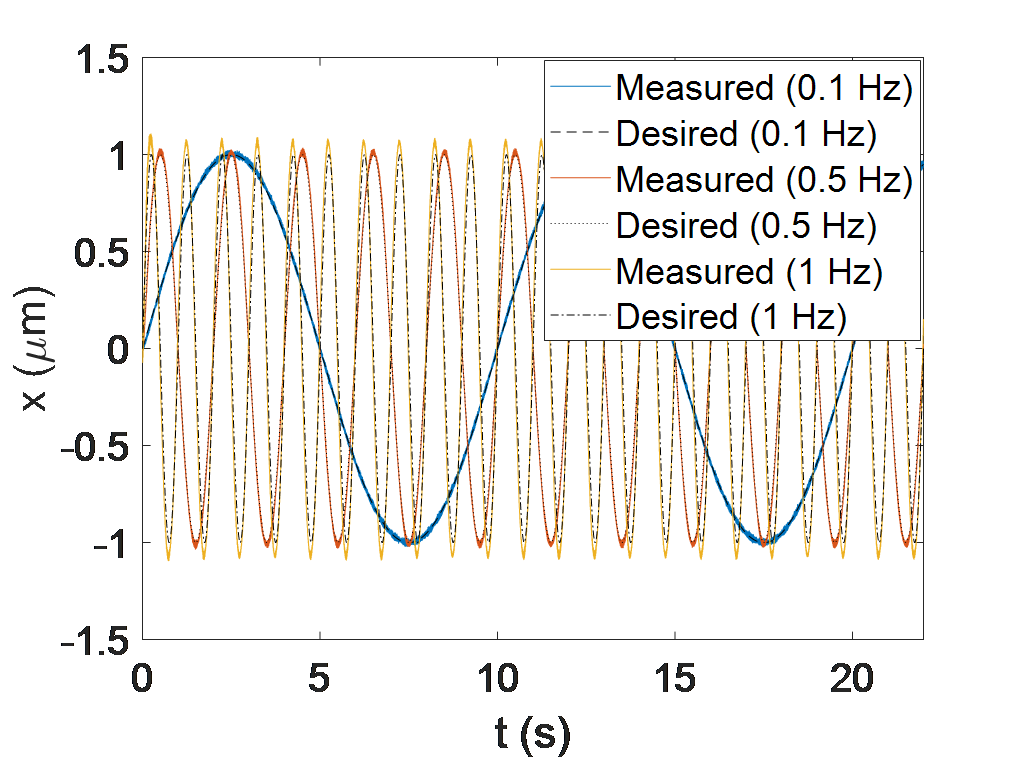}} & \hbox{\hspace{-2.5em}\includegraphics[width=5cm]{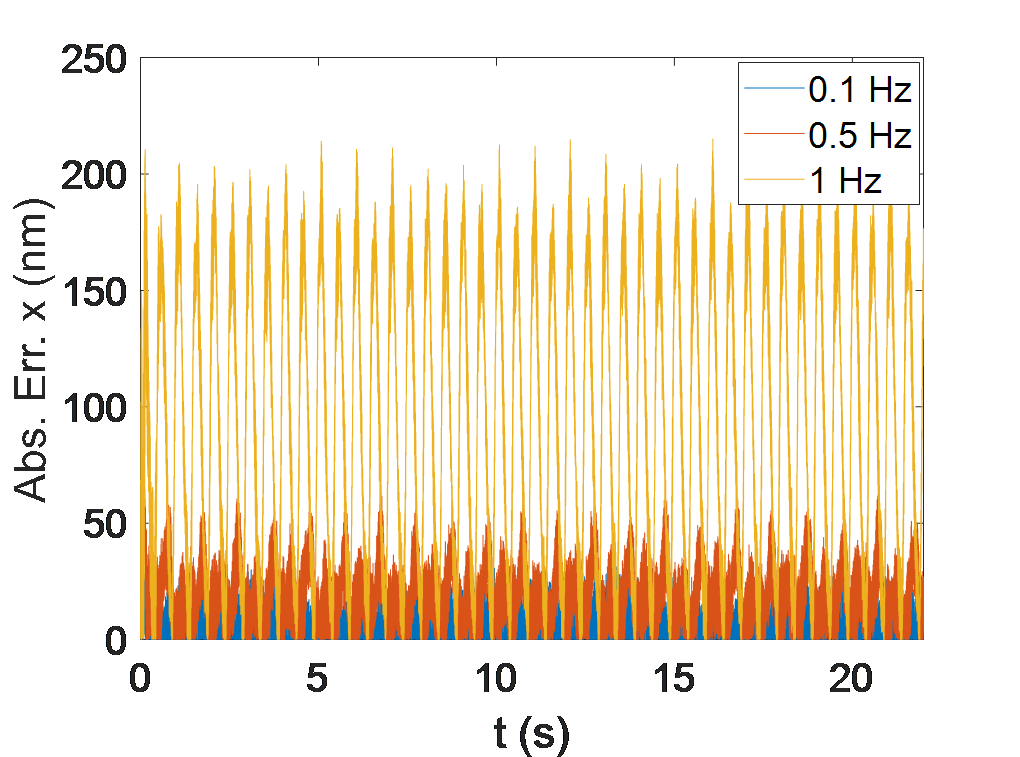}} \\ [0ex]
        \hbox{\hspace{-1.0em}\includegraphics[width=5cm]{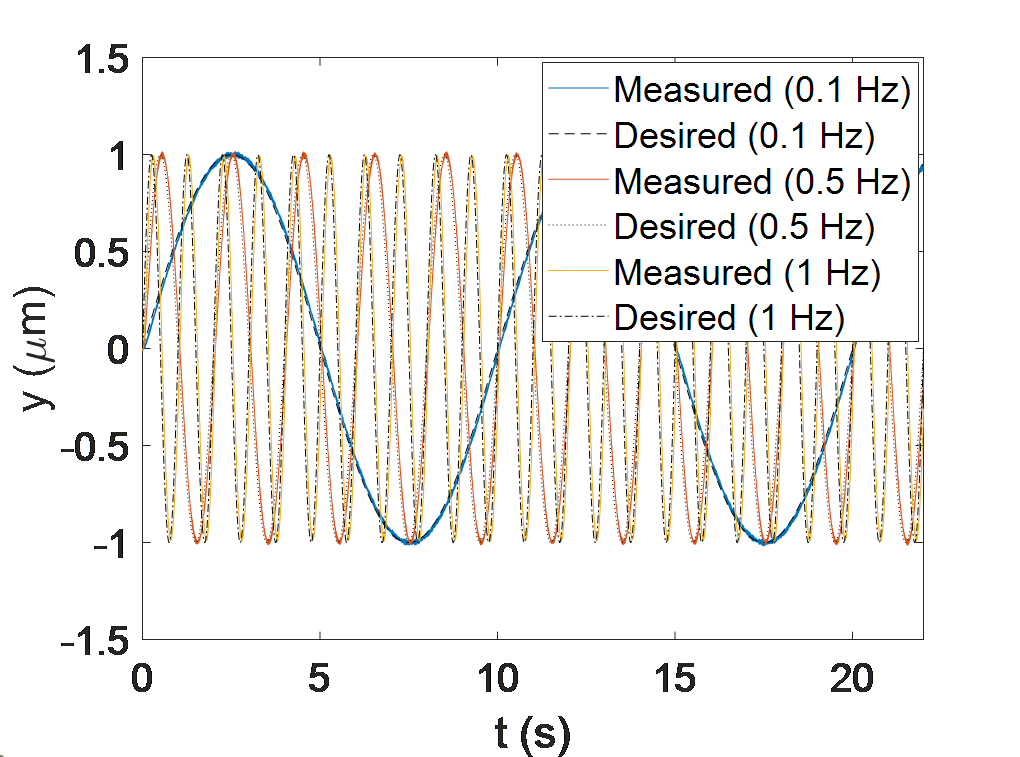}} & \hbox{\hspace{-2.5em}\includegraphics[width=5cm]{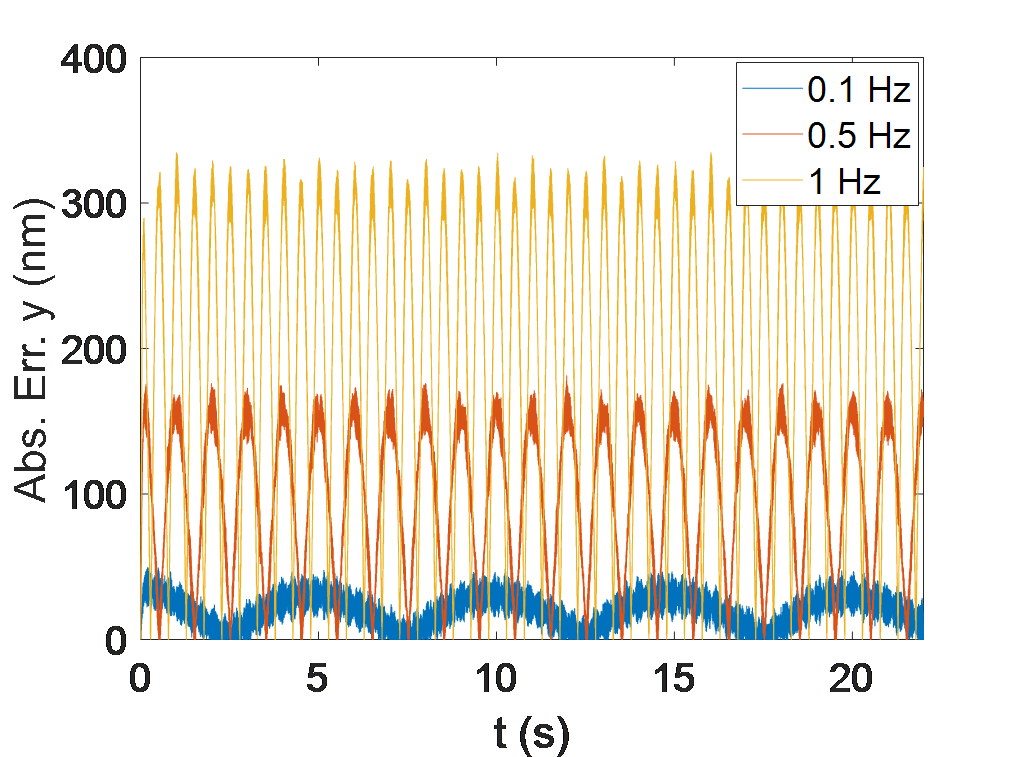}} \\ [0ex]
        \hbox{\hspace{-1.0em}\includegraphics[width=5cm]{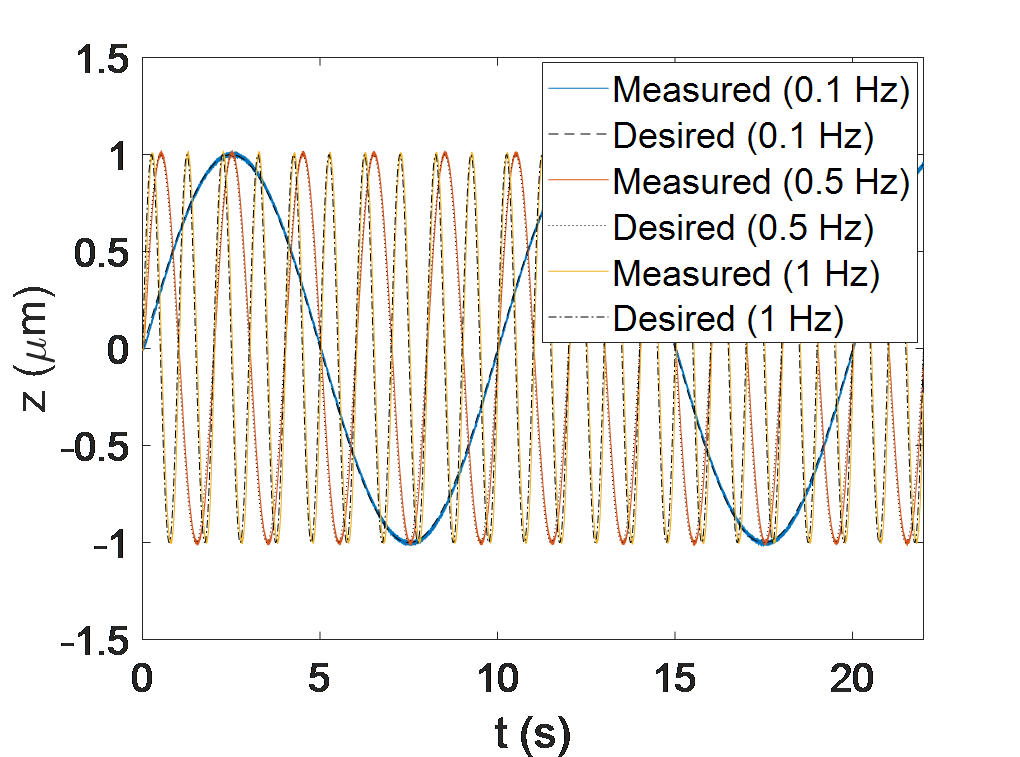}} & \hbox{\hspace{-2.5em}\includegraphics[width=5cm]{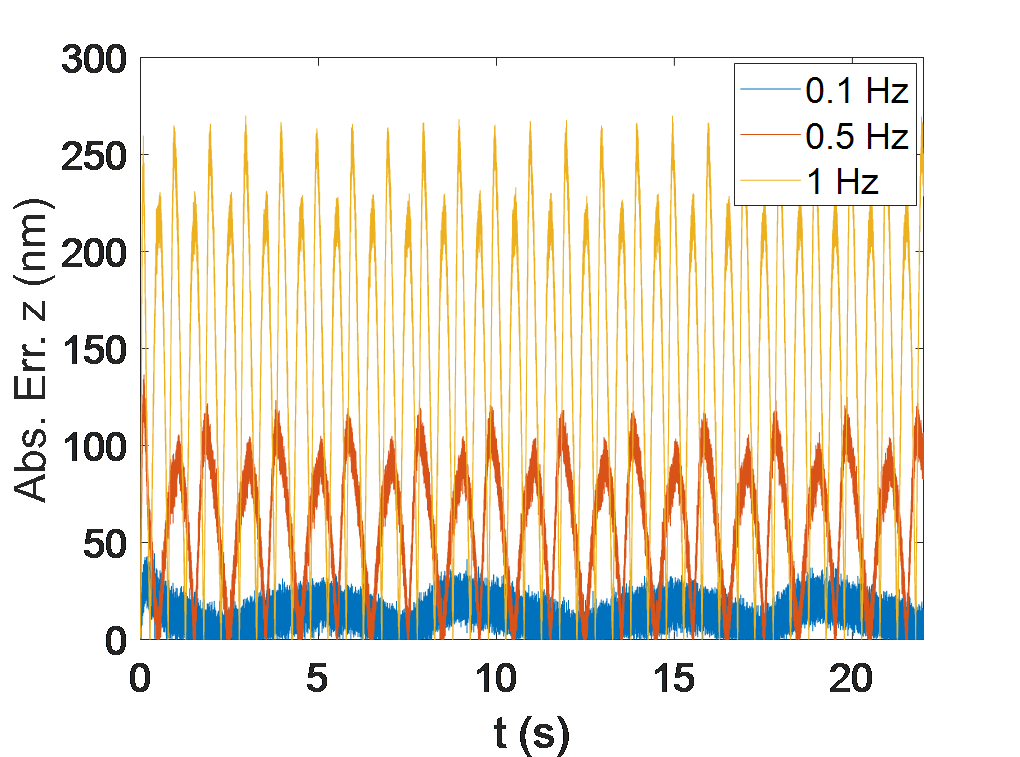}} \\ [0ex]
        \hbox{\hspace{-1.0em}\includegraphics[width=5cm]{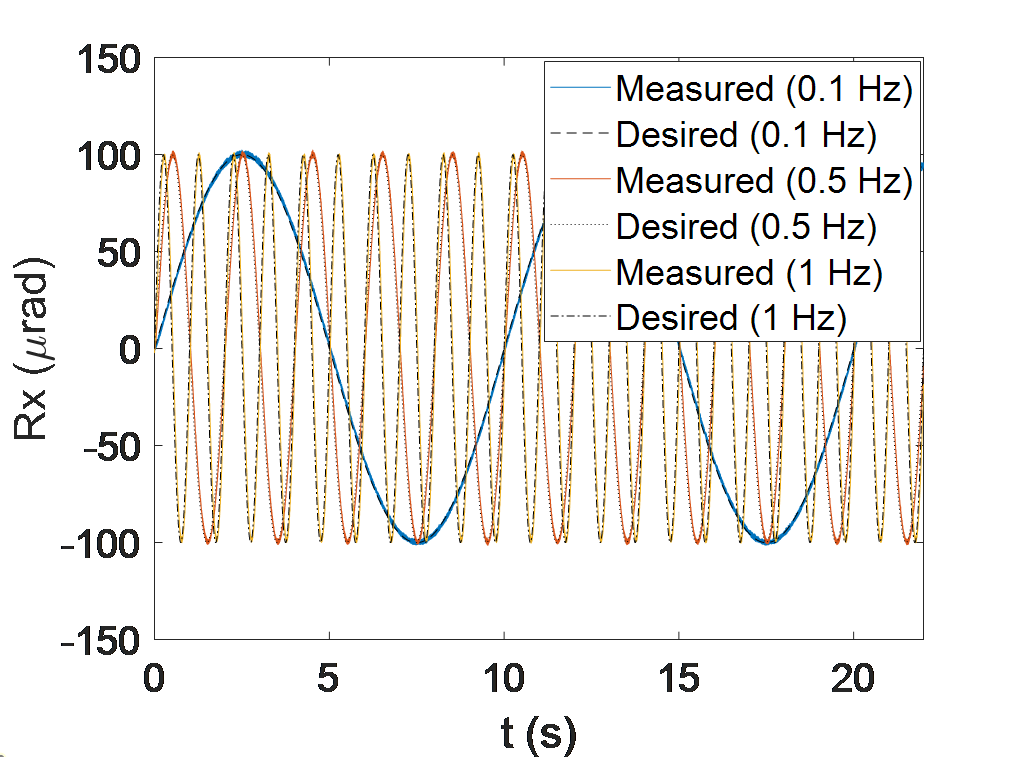}} & \hbox{\hspace{-2.5em}\includegraphics[width=5cm]{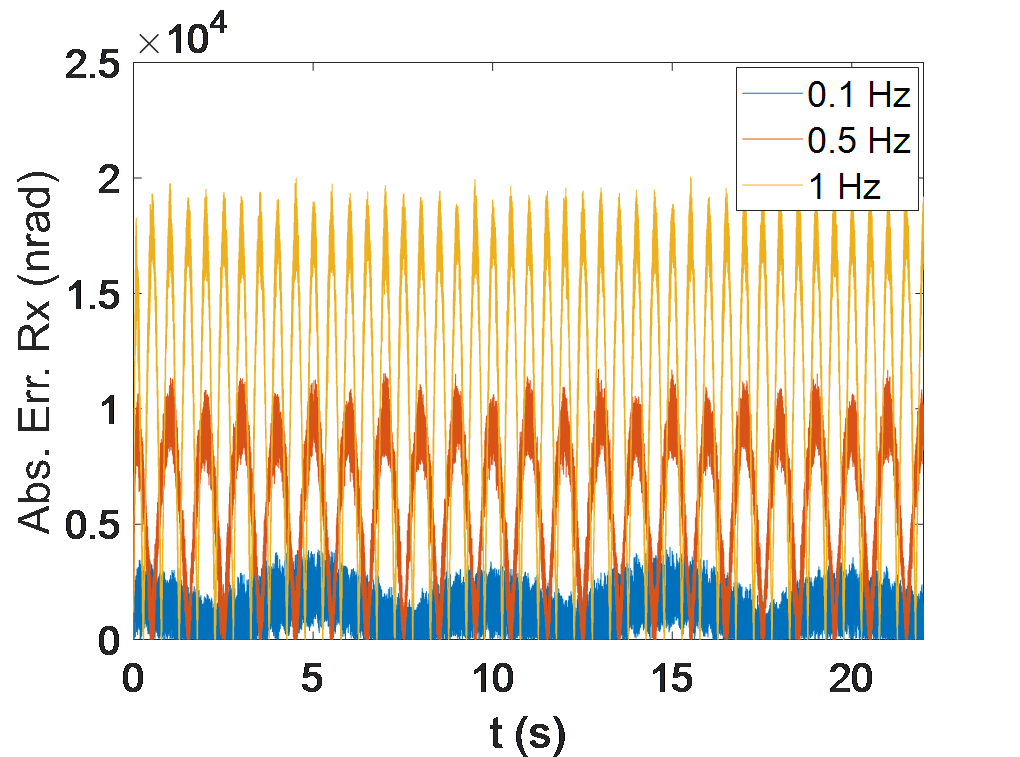}} \\ [0ex]
        \hbox{\hspace{-1.0em}\includegraphics[width=5cm]{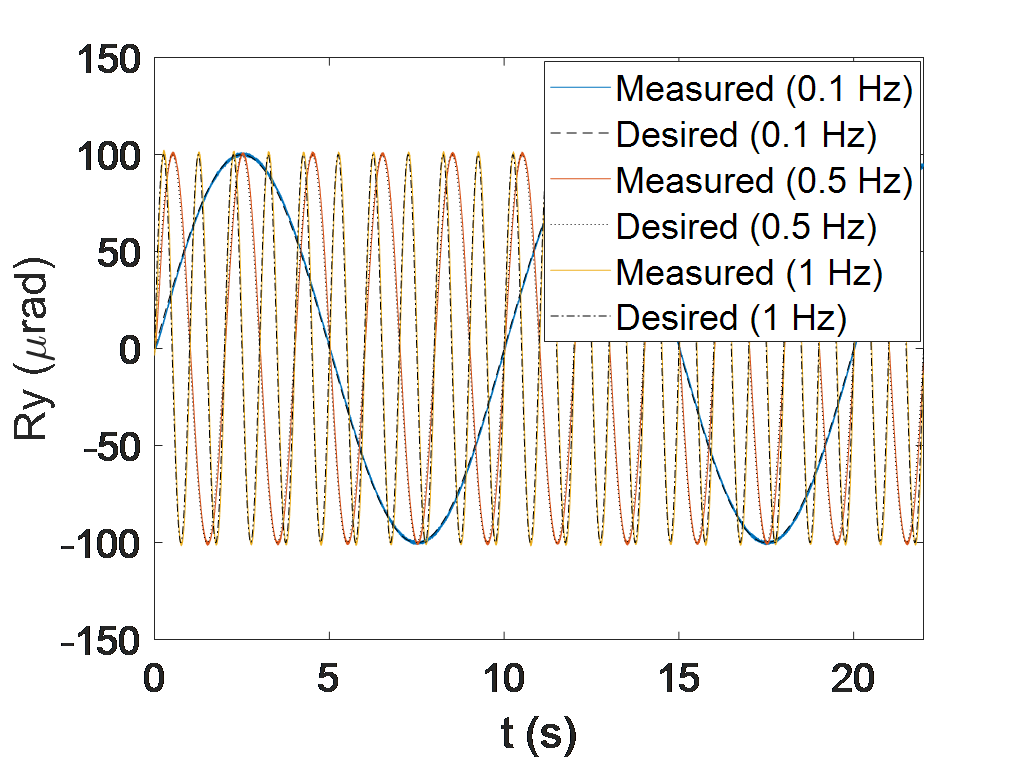}} & \hbox{\hspace{-2.5em}\includegraphics[width=5cm]{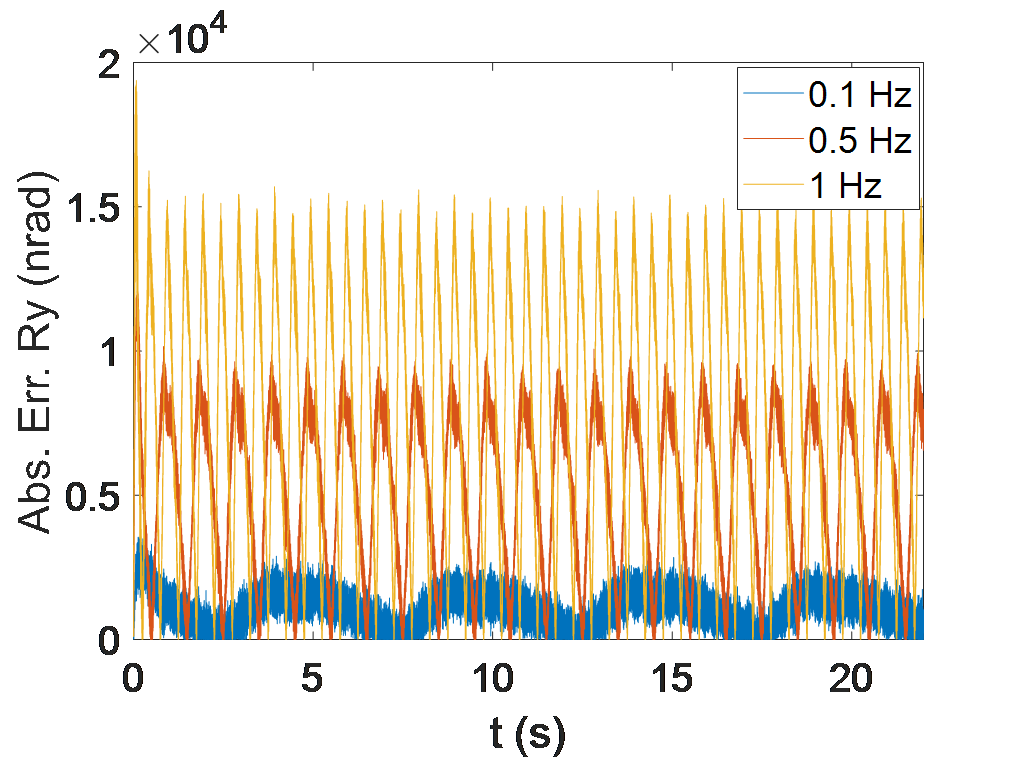}} \\ [0ex]
        \hbox{\hspace{-1.0em}\includegraphics[width=5cm]{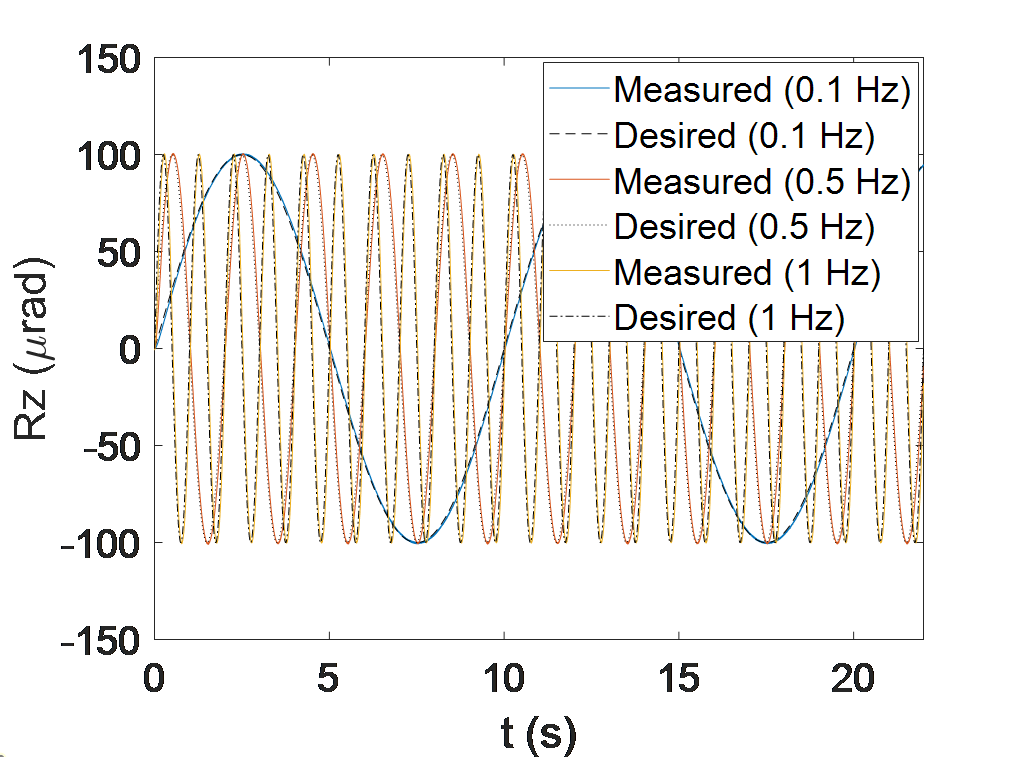}} & \hbox{\hspace{-2.5em}\includegraphics[width=5cm]{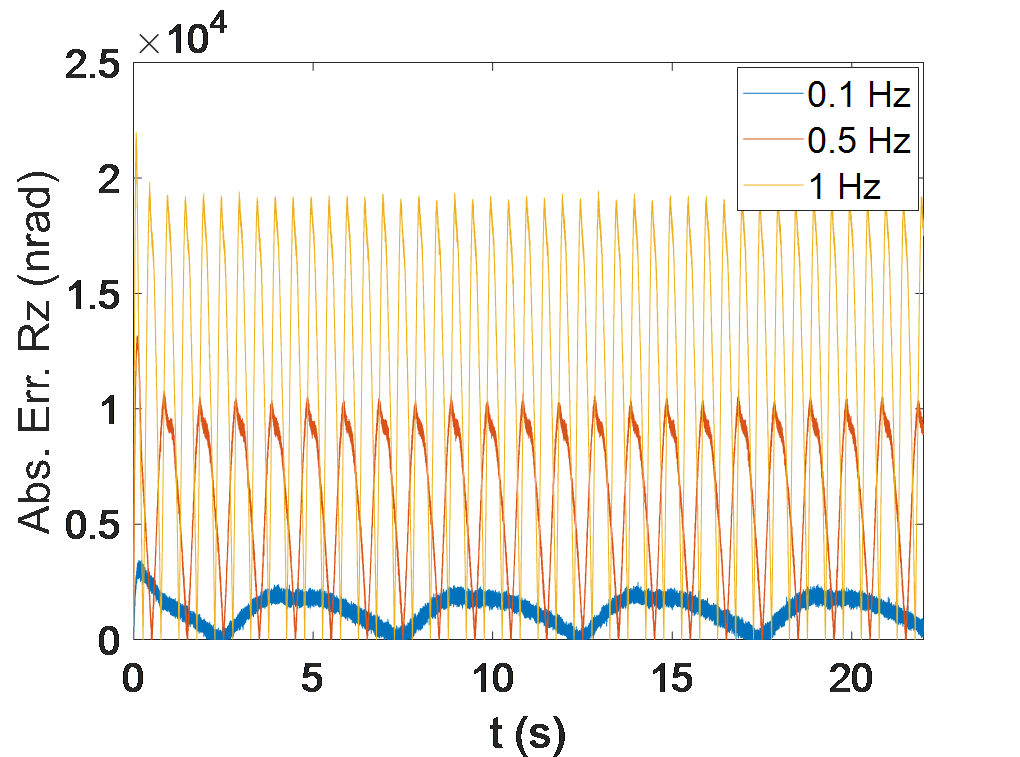}} \\ [0ex]
	\end{tabular}
\caption{Effect of manipulation frequency on the accuracy of the manipulation task}
\label{fig_11}
\end{figure}

\noindent The hysteresis effect inherent in piezoelectric ceramics actuators is an undesired phenomenon that affects the performances of precise positioning systems. To investigate the capability of the implemented feedback control scheme for hysteresis reduction, a series of sinusoidal inputs were generated and applied to the proposed piezo-actuated positioner. The results are shown in Fig. 12. Considering the obtained results, it can be observed the relationship between the input and output of the system in each translational and rotational direction was linear which can validate that the hysteresis phenomenon was reduced significantly.

\begin{figure*}[htbp]\centering
\begin{tabular}{l l l}
        \hbox{\hspace{-1.0em}\includegraphics[width=4.5cm]{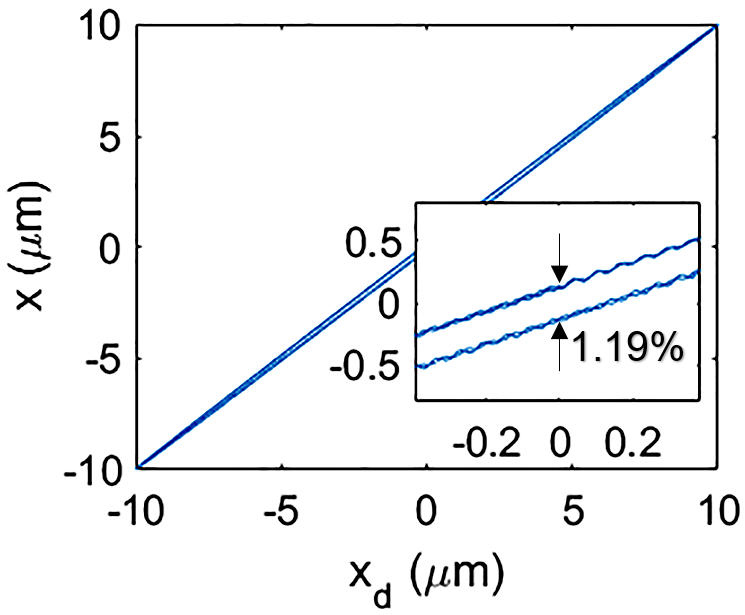}} & \hbox{\hspace{-1.0em}\includegraphics[width=4.5cm]{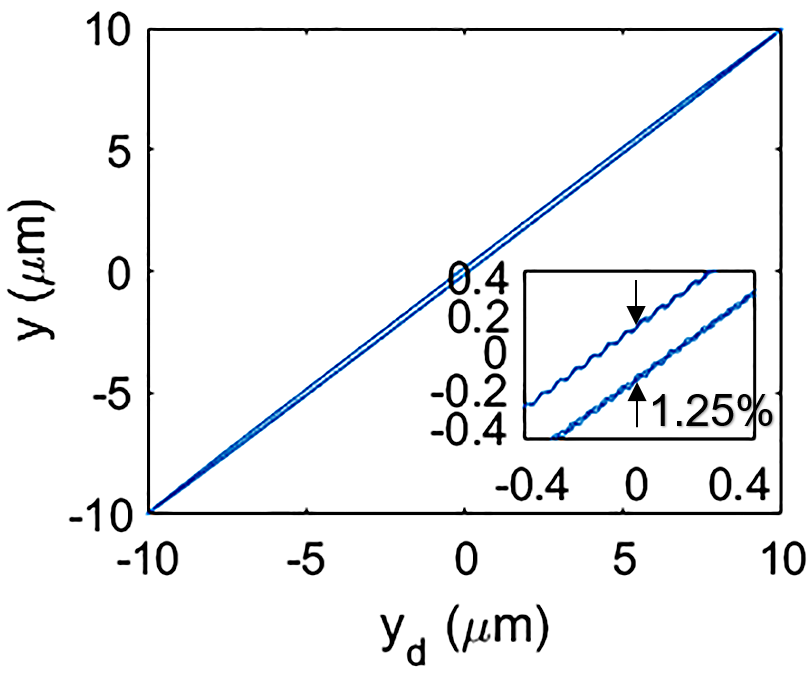}} & \hbox{\hspace{-1.0em}\includegraphics[width=4.5cm]{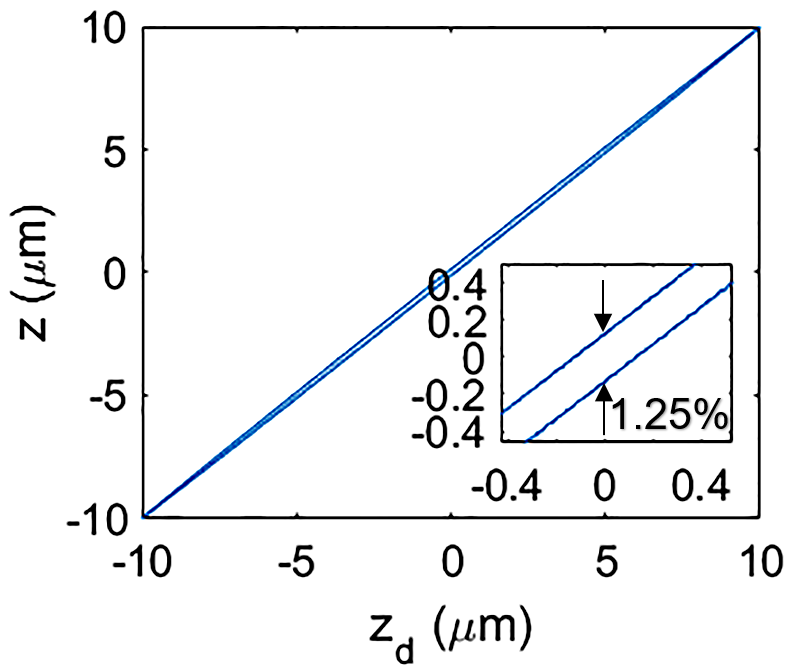}} \\ [0ex]
        \hbox{\hspace{-1.0em}\includegraphics[width=4.5cm]{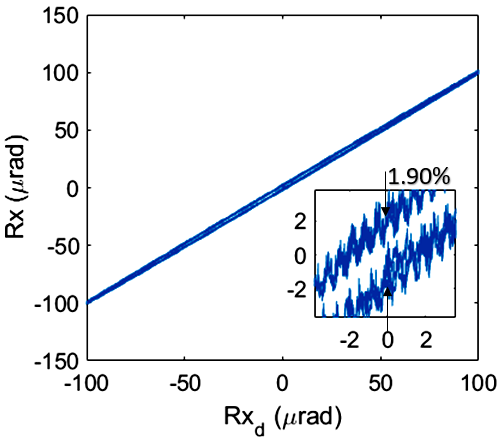}} & \hbox{\hspace{-1.0em}\includegraphics[width=4.5cm]{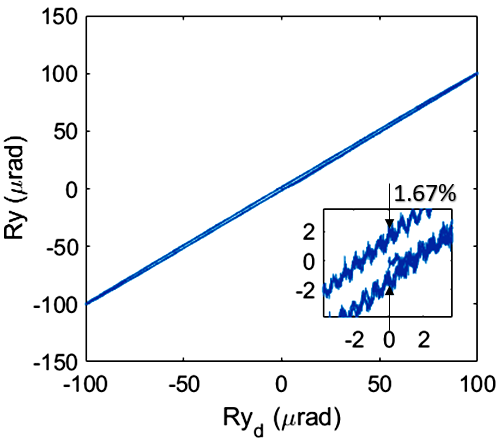}} & \hbox{\hspace{-1.0em}\includegraphics[width=4.5cm]{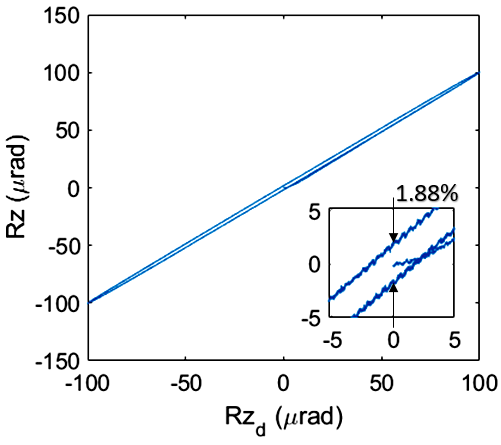}} \\ [0ex]
	\end{tabular}
\caption{Actual-reference hysteresis compensated loops of the six-DOF piezo-actuated positioner}
\label{fig_12}
\vspace{-5mm}
\end{figure*}

\vspace{-1mm}

\subsection{Frequency Response Experiment}\label{D3}
A frequency analysis test was conducted to investigate the dynamic characteristics of the proposed six-DOF positioner. Fig. 13 shows the experimental results of the first six frequency mode test. To capture these results, a series of sinusoidal sweep signals with their frequency increasing linearly was applied to the PEAs as the inputs of the system. Data for this experiment were gathered at a control frequency of 10kHz. The results demonstrated reasonable predictions of the resonant modes in comparison with the computational results. The inherent resonant frequencies are higher than the calculated ones due to the carrying mass. Moreover, using stiffer material in the manufacturing process will result in a positioner with higher bandwidth frequencies for high-speed real-time applications.

\begin{figure}[t]\centering
\begin{tabular}{l l}
        \hbox{\hspace{-2.0em}\includegraphics[width=5cm]{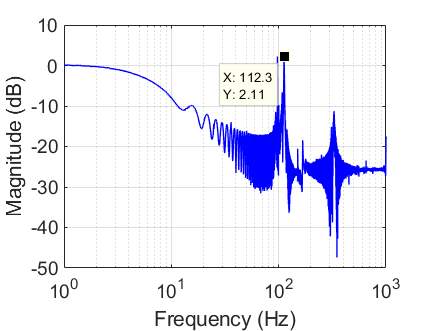}} & \hbox{\hspace{-2.0em}\includegraphics[width=5cm]{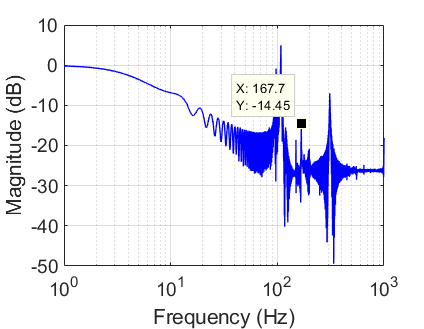}} \\ [0ex]
        \hbox{\hspace{-2.0em}\includegraphics[width=5cm]{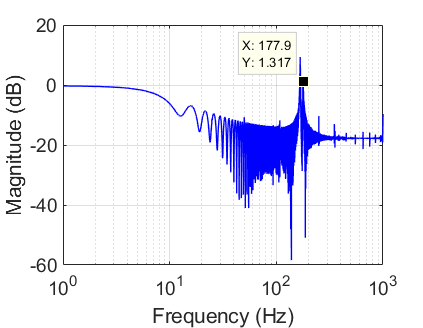}} & \hbox{\hspace{-2.0em}\includegraphics[width=5cm]{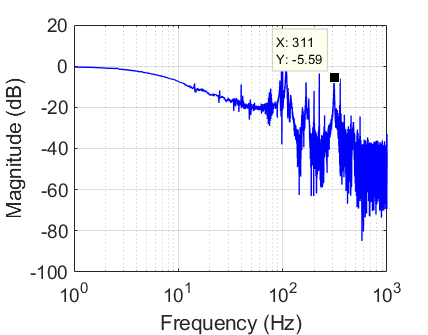}} \\ [0ex]
        \hbox{\hspace{-2.0em}\includegraphics[width=5cm]{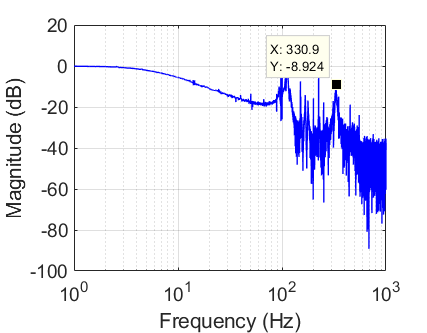}} & \hbox{\hspace{-2.0em}\includegraphics[width=5cm]{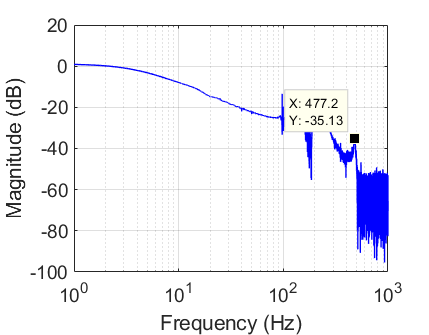}} \\ [0ex]
	\end{tabular}
\caption{Results of frequency analysis of the six-DOF positioner}
\label{fig_13}
\vspace{-5mm}
\end{figure}

\vspace{-1mm}

\section{Discussion}\label{E}
A compliant parallel monolithic large range six-DOF precise positioning system was presented in this work. The system was manufactured as a single consistent monolithic structure using a 3D printing technique. Extensive computational and experimental evaluations of the performances of the positioner were shown to provide an in-depth understanding of its capabilities and characteristics. Due to the large range, high resolution, and compactness of the positioner, it can be modified to be utilized in many applications including pick-and-place manipulation, tremor compensation in microsurgery and micro-assembly, and collaborative manipulators systems. In order to improve the accuracy of positioning tasks performed by the proposed positioner, a robust adaptive \cite{b33} disturbance observer-based controller can be established \cite{b34,b35,b36}. In addition, uncertainty analysis can be conducted to improve the accuracy of rotational measurements.

\vspace{-1mm}



\vspace{-15mm}

\begin{IEEEbiography}[{\includegraphics[width=1in,height=1.25in,clip,keepaspectratio]{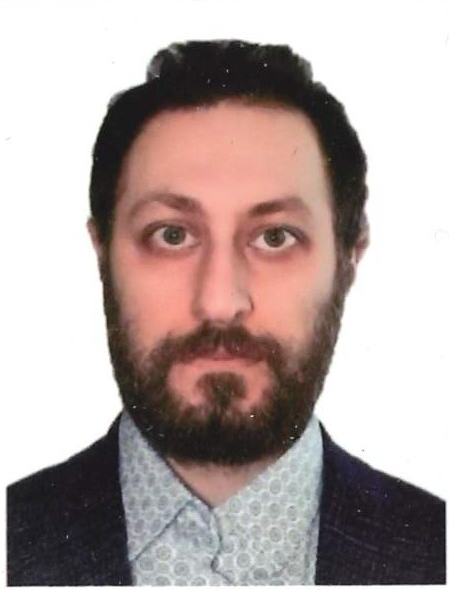}}]{Mohammadali Ghafarian} (Member, IEEE) was the member of the Robotics and Mechatronics Research Laboratory (RMRL), and received his Ph.D. degree in 2021 in the field of robotics and mechatronics from Monash University. He is a member of the Institute of Electrical and Electronics Engineers (IEEE), and a Member of the Institution of Engineers Australia (MIEAust). He is currently working on establishing the next-generation of vehicular motion simulator at the Institute for Intelligent Systems Research and Innovation (IISRI), Deakin University, Australia. His current research interests include robotic systems, robust/adaptive control system design, real-time systems, sensor fusion, and finite element modelling.
\end{IEEEbiography}

\vspace{-10mm}

\begin{IEEEbiography}[{\includegraphics[width=1in,height=1.25in,clip,keepaspectratio]{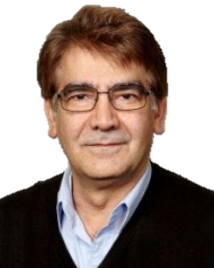}}]{Bijan Shirinzadeh} received the B.Eng. degree in mechanical engineering, the B.Eng. degree in aerospace engineering, and the M.S.Eng. degree in mechanical engineering from the University of Michigan, Ann Arbor, and the Ph.D. degree in mechanical engineering from The University of Western Australia (UWA). He is currently a Full Professor and the Director of the Robotics and Mechatronics Research Laboratory (RMRL). His current
research interests include laser-based measurements and sensory-based control, micro/nanomanipulation mechanisms and systems, serial and parallel mechanisms, automated manufacturing, and robotic surgery.
\end{IEEEbiography}

\vspace{-5mm}

\begin{IEEEbiography}[{\includegraphics[width=1in,height=1.25in,clip,keepaspectratio]{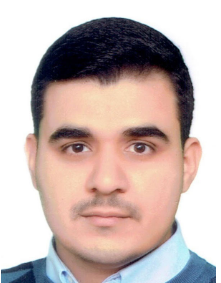}}]{Ammar Al-Jodah} received the B.Sc. degree in control and systems engineering from the University of Technology, Iraq, and the M.Sc. degree in electrical and computer engineering from Southern Illinois University, Carbondale, USA. He finished his Ph.D. degree with the Robotics and Mechatronics Research Laboratory (RMRL), Monash University, Australia. His research interests include large range micro/nanomanipulation mechanisms, laser interferometry-based measurement systems, robust controller design, and real-time controllers.
\end{IEEEbiography}

\vspace{0mm}

\vfill

\end{document}